\documentclass{article}

    \PassOptionsToPackage{numbers, compress}{natbib}

 \usepackage[preprint]{neurips_2026}

\usepackage[utf8]{inputenc}
\usepackage[T1]{fontenc}
\usepackage{hyperref}
\usepackage{url}
\usepackage{booktabs}
\usepackage{amsfonts}
\usepackage{nicefrac}
\usepackage{microtype}
\usepackage{xcolor}

\usepackage{amsmath}
\usepackage{wrapfig}
\usepackage{xspace}
\newcommand{\modelname}{Lip Forcing\xspace}
\usepackage{graphicx}
\usepackage{multirow}
\usepackage{lipsum}
\usepackage{enumitem}
\usepackage{algorithm}
\usepackage{algpseudocode}
\usepackage{tikz}
\usepackage{setspace}

\newcommand{\supp}[1]{App.~\ref{#1}}

\newcommand{\paragrapht}[1]{\noindent\textbf{#1}}

\title{Lip Forcing: Few-Step Autoregressive Diffusion\\ for Real-time Lip Synchronization}

\author{
\textbf{Paul Hyunbin Cho}$^{1\ast}$ 
\quad \textbf{Jinhyuk Jang}$^{1\ast}$ 
\quad \textbf{SeokYoung Lee}$^1$ 
\\
\textbf{Joungbin Lee}$^1$ 
\quad \textbf{Siyoon Jin}$^1$ 
\quad \textbf{Heeseong Shin}$^1$
\quad \textbf{Jung Yi}$^1$ 
\\
\textbf{Yunjin Park}$^2$ 
\quad \textbf{Chulmin Park}$^{2}$  
\quad \textbf{Seungryong Kim}$^{1\dagger}$ 
\\[0.5em]
{
$^1$KAIST AI 
\qquad $^2$AIPARK 
}
\\[0.5em]
{\tt \href{https://cvlab-kaist.github.io/LipForcing/}{\textcolor{purple}{https://cvlab-kaist.github.io/LipForcing}}}
}

\begin{document}
\begingroup
\renewcommand{\thefootnote}{}
\footnotetext{$^\ast$: Equal contribution}
\footnotetext{$^\dagger$: Corresponding authors}
\endgroup

\maketitle

\begin{figure}[h]
    \centering
   \vspace{-10pt}
   \includegraphics[width=\linewidth]{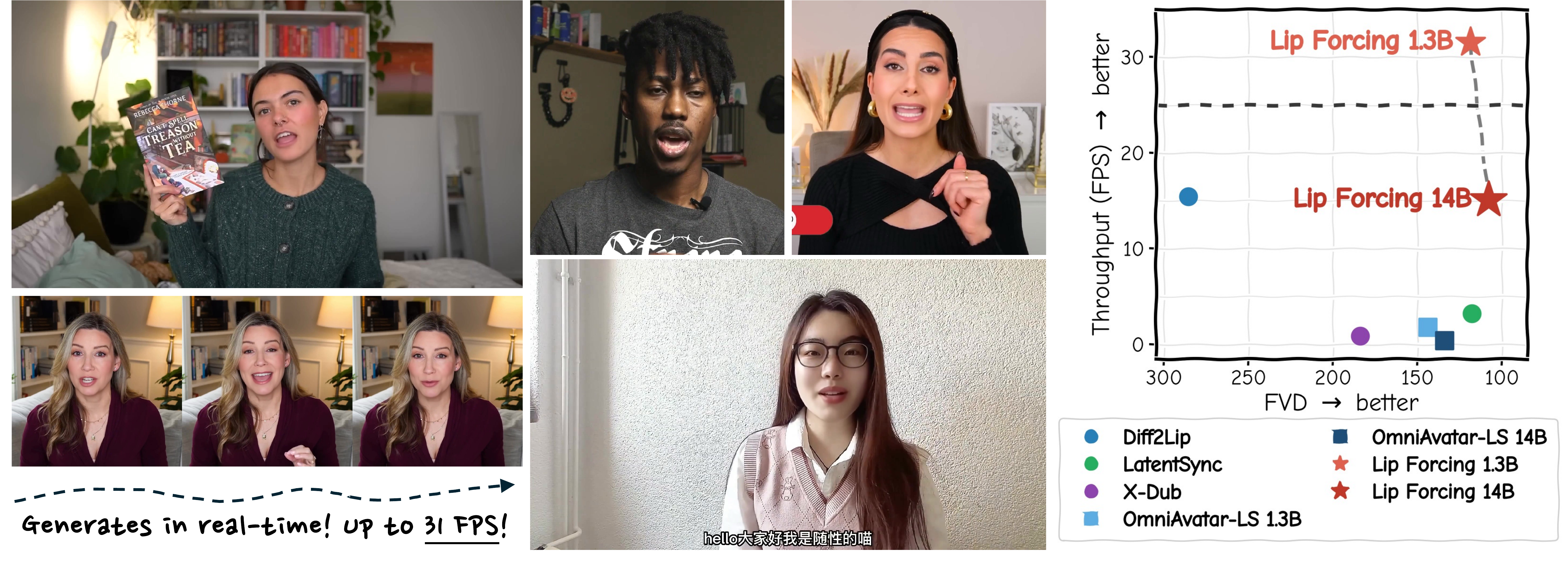}
    \vspace{-15pt}
    \caption{
    \textbf{\modelname.} A streaming model for real-time lip synchronization that produces photorealistic, accurately lip-synced video at up to \textbf{31 FPS} with low latency and memory. 
    \emph{Right:} both student scales lie on the throughput--FVD Pareto frontier, ahead of prior diffusion lip-sync methods.
    }
    \label{fig:teaser}
\end{figure}

\begin{abstract}
Diffusion-based lip synchronization models achieve strong visual quality and audio-visual alignment, but full-sequence bidirectional attention and many denoising steps make them impractical for real-time inference.
We present \modelname, to our knowledge the first autoregressive diffusion method for video-to-video (V2V) lip synchronization, which distills a 14B audio-conditioned bidirectional video diffusion teacher into causal students.
At inference, the students generate each chunk in only \textbf{two} denoising steps without inference-time CFG, enabling real-time lip synchronization. 
A lip-sync-specific teacher-trajectory analysis reveals a \textit{CFG fidelity--sync tradeoff}: no-CFG predictions favor reference fidelity, whereas CFG-guided predictions favor synchronization within a mid-trajectory band.
\modelname translates this finding into three analysis-derived components: Sync-Window DMD, a two-step inference schedule, and a SyncNet-based reward.
We validate \modelname at two student scales, both distilled from the 14B teacher.
The 1.3B student crosses into real-time streaming at 31 FPS, \textbf{17.6×} faster than its same-scale bidirectional model.
The 14B student, the largest diffusion model reported for V2V lip synchronization, runs \textbf{39.8×} faster than its teacher at comparable reference fidelity.
Time-to-first-frame is sub-millisecond at both scales, far below every diffusion baseline.

\end{abstract}

\section{Introduction}
Audio-driven video-to-video (V2V) lip synchronization~\cite{Prajwal_2020, mukhopadhyay2023diff2lipaudioconditioneddiffusion, cheng2022videoretalkingaudiobasedlipsynchronization, zhang2025musetalkrealtimehighfidelityvideo} aims to synthesize mouth motion in a source video that matches a target audio signal while preserving the speaker's identity, head pose, and background.
Recent diffusion-based methods~\cite{li2024latentsync, peng2025omnisyncuniversallipsynchronization, wan2025} have substantially improved visual fidelity and audio-visual alignment. However, their high inference cost limits their use in practical deployment scenarios, from offline dubbing to latency-sensitive applications such as live translation, virtual avatars, and interactive agents.
Several approaches have sought to reduce this cost~\cite{zhang2025musetalkrealtimehighfidelityvideo}, but often at the expense of visual fidelity and realism.

This deployment gap stems from two main computational factors. First, recent transformer-based diffusion models compute self-attention over the entire video sequence~\cite{li2024latentsync, peng2025omnisyncuniversallipsynchronization, esser2024scalingrectifiedflowtransformers}, scaling quadratically with clip length. 
Although autoregressive diffusion methods~\cite{chen2024diffusion, ai2025magi1autoregressivevideogeneration, huang2025self, yin2025slow} can alleviate this burden through chunk-wise causal generation, they remain largely unexplored for lip synchronization. 
Second, achieving high-quality synthesis in existing frameworks typically requires tens of denoising steps, significantly compounding the overall computational cost. 
Meanwhile, few-step distillation methods~\cite{huang2025self}, which are commonly employed to reduce denoising steps, are underexplored in the context of lip synchronization. 
We argue that integrating such methods requires careful consideration as na\"ive adaptations can easily introduce unexpected artifacts due to the complex entanglement of audio-visual signals in lip synchronization settings.

In this paper, we propose \textbf{\modelname}, a lip-sync-specialized distillation framework that compresses a 50-step bidirectional teacher~\cite{gan2025omniavatar} into a \textit{two-step} streaming student. 
To make few-step distillation work for lip synchronization rather than degrade it, we first analyze a bidirectional lip-sync diffusion model (Sec.~\ref{sec:method_analysis}) and identify a \emph{CFG fidelity--sync tradeoff} inherent to the denoising process, which we then exploit during distillation and inference.
Specifically, we observe that the model better preserves reference fidelity without classifier-free guidance (CFG), whereas applying CFG significantly improves audio-visual synchronization mainly within a mid-trajectory band. 
This suggests that different timesteps throughout the denoising trajectory exhibit varying degrees of responsiveness to audio conditioning; consequently, employing a single, fixed guidance scale can lead to a suboptimal compromise between identity preservation and accurate lip movements.
We then run a no-CFG$\to$CFG-guided Euler-step probe to locate a two-step operating point near a mid-trajectory landing, as shown in Fig.~\ref{fig:analysis_combined}.
Together, these analyses determine the training guidance window and the student's landing step.

We instantiate the training guidance window as \textbf{Sync-Window DMD (SW-DMD)}, which
replaces standard DMD's fixed teacher guidance with a sync-window guidance schedule that enables CFG only on training timesteps inside the sync-favoring band identified by the analysis.
At inference, the student follows a \textbf{two-step inference schedule}, denoising in two model calls with the second placed at the analysis-derived landing point.
A \textbf{SyncNet-based reward} adds explicit lip-sync supervision to the distillation objective.

We distill a 14B teacher into students at two scales, 1.3B and 14B, and evaluate
on HDTF~\cite{zhang2021flow},
where \modelname improves the streaming speed--fidelity tradeoff of diffusion-based
V2V lip synchronization and enables streaming deployment that bidirectional methods
cannot serve.

In summary, our contributions are:
\begin{itemize}
    \item We propose \textbf{\modelname}, an analysis-driven distillation framework and, to our knowledge, the \textbf{first autoregressive diffusion method for V2V lip synchronization}, enabling real-time deployment
    with causal autoregressive students.

    \item We provide a \emph{lip-sync-specific teacher-trajectory analysis} that characterizes the teacher's denoising behavior and motivates a three-part distillation recipe: Sync-Window DMD (SW-DMD), a two-step inference schedule, and a SyncNet-based reward.

    \item We validate \modelname at two student scales. The 1.3B student reaches {31 FPS} ($17.6{\times}$ faster than its same-scale bidirectional model), crossing the 25~FPS real-time threshold, while the 14B student, the largest diffusion student reported to date for V2V lip synchronization, runs $39.8{\times}$ faster than its teacher and $4.7{\times}$ faster than LatentSync at comparable reference fidelity. Time-to-first-frame (TTFF) is sub-millisecond at both scales.
\end{itemize}

\section{Related Work}

\subsection{Audio-driven lip synchronization}
Audio-driven lip synchronization has been explored through image-to-video (I2V) portrait animation~\cite{cui2025hallo3highlydynamicrealistic, wang2025fantasytalkingrealistictalkingportrait,gan2025omniavatar} that generates talking faces from a static image, and video-to-video (V2V) lip sync~\cite{Prajwal_2020, ma2025sayanythingaudiodrivenlipsynchronization, mukhopadhyay2023diff2lipaudioconditioneddiffusion, cheng2022videoretalkingaudiobasedlipsynchronization} that edits an existing video to match target audio while preserving identity, pose, and background; we focus on the latter.
Early V2V methods~\cite{Prajwal_2020, wang2023seeingsaidtalkingface} used GANs~\cite{goodfellow2014generativeadversarialnetworks} for efficient inference but suffered from blur and temporal inconsistency.
Following progress in diffusion-based generation~\cite{ho2020denoising, song2022denoisingdiffusionimplicitmodels, rombach2022high, wan2025, yang2024cogvideox, kong2024hunyuanvideo} and its video applications~\cite{lee20253dscenepromptingsceneconsistent, lee2025vwarperappearanceconsistentvideodiffusion, jin2025matrixmasktrackalignment}, recent lip-sync methods adopt diffusion backbones~\cite{cheng2022videoretalkingaudiobasedlipsynchronization, mukhopadhyay2023diff2lipaudioconditioneddiffusion, zhang2025musetalkrealtimehighfidelityvideo, kim2024moditalkermotiondisentangleddiffusionmodel, li2024latentsync, he2025inpainting} for improved quality and alignment, but their bidirectional full-context attention~\cite{esser2024scalingrectifiedflowtransformers} scales poorly with sequence length.
We instead propose an autoregressive diffusion transformer for lip synchronization that generates frames sequentially conditioned on past frames.

\subsection{Autoregressive video diffusion models}
Autoregressive video diffusion models~\cite{chen2024diffusion, ai2025magi1autoregressivevideogeneration, yin2025slow, huang2025self} generate frames or chunks sequentially with KV caching, enabling streaming inference.
Recent work distills bidirectional teachers into causal few-step students via Distribution Matching Distillation (DMD)~\cite{yin2024one, yin2024improved}, with Self Forcing~\cite{huang2025self} additionally training on self-rollouts to remove the train-test exposure mismatch. Extensions add sink-frames, rolling-reference attention~\cite{yang2025longlive, yi2025deep, liu2025rollingforcingautoregressivelong}, or related mechanisms to stabilize extrapolation, while reward-weighted variants~\cite{lu2025reward} reweight the per-sample DMD gradient by a task-specific reward to improve dynamics.
Recent works have also distilled task-conditional teachers into causal streaming students: MotionStream~\cite{shin2025motionstreamrealtimevideogeneration} for trajectory-conditioned synthesis and Live Avatar~\cite{huang2025liveavatarstreamingrealtime} for audio-driven avatars, specializing the conditioning architecture or inference system while inheriting the underlying distillation recipe unchanged.
We instead specialize the distillation objective itself for lip synchronization, derived from a trajectory-level analysis of the teacher.

\section{Preliminaries}
\label{sec:preliminaries}
\paragrapht{Rectified flow.}
Rectified flow~\cite{liu2022flow} parameterizes diffusion-based generation as a deterministic transport process from noise to data, simplifying sampling through velocity-based updates.
Given a real sample $x_0 \sim p_0(x)$ and Gaussian noise $\epsilon \sim \mathcal{N}(0,I)$, the intermediate state at timestep $t \in [0,1]$ is defined as:
\begin{equation}
    \label{eq:rf_interpolation}
    x_t = (1 - t)\,x_0 + t\,\epsilon.
\end{equation}
The model is trained to predict a velocity field $v_\theta(\cdot)$ that deterministically transports samples along the interpolation path via a flow-matching objective.
Given the current state $x_t$, the rectified flow model allows estimating an earlier state $\hat{x}_{t-\Delta t}$ by applying a deterministic backward update using the predicted velocity field:
\begin{equation}
    \label{eq:psi}
    \hat{x}_{t-\Delta t}
    = \Psi(x_t, v_\theta(x_t, t),t-\Delta t)
    = x_t - \Delta t~ v_\theta(x_t, t) ,
\end{equation}
where $\Psi(\cdot)$ denotes a deterministic backward flow operator.

As a special case, when $\Delta t = t$, the original clean sample $x_0$
can be directly estimated from $x_t$ as:
\begin{equation}
\label{eq:x0_from_v}
\hat{x}_0 = \Psi(x_t, v_\theta(x_t,t), 0)
= x_t - t\, v_\theta(x_t,t).
\end{equation}
We use \(t\in[0,1]\) for continuous rectified-flow time, with \(t=1\) denoting noise and \(t=0\) denoting data. 
The teacher sampler uses a fixed shifted ODE schedule of 50 steps over 51 nodes \(\{\tau_j\}_{j=0}^{50}\), ordered from noisy (\(\tau_0=0.999\)) to clean (\(\tau_{50}=0\)); the model is evaluated at the 50 step indices \(j=0,\ldots,49\), and the final step lands on the clean sample \(\tau_{50}=0\). 
We use \(j\) only for this discrete ODE step index and \(\tau_j\) for the corresponding continuous timestep; for example, \(\tau_0=0.999\) and \(\tau_{30}=0.769\).
For simplicity, we omit the VAE notation in equations. 

\paragrapht{Self Forcing and DMD.}
To enable causal streaming generation, autoregressive video diffusion models generate each frame or chunk conditioned only on previously generated clean outputs. 
For frame/chunk \(i\), a causal student \(G_\theta\) predicts a rectified-flow velocity at scheduled time \(\tau_j\), conditioned on its own previous clean predictions \(\hat{x}^{<i}_0\) and conditioning inputs \(c\). 
A \(K\)-call student uses a subset of teacher ODE indices \(J=(j_0,\ldots,j_{K-1})\), with \(K\ll 50\), and repeatedly applies the backward flow operator \(\Psi\) before projecting the final state to \(\hat{x}_0\).

We adopt Self Forcing on top of DMD. Given the student clean prediction \(\hat{x}_0\), DMD re-noises it as \(x_t=(1-t)\hat{x}_0+t\epsilon\) where \(q(t)\) is a  distribution over \([0,1]\) and \(\epsilon \sim \mathcal{N}(0,I)\), and updates the student using the difference between a frozen teacher score and a learned fake-score network:
\begin{equation}
\label{eq:dmd_update}
\nabla_\theta \mathcal{L}_{\mathrm{DMD}}
\propto
\mathbb{E}_{t,\epsilon}
\left[
\left(
S^{\mathrm{CFG}}_{\mathrm{real}}(x_t,t,c;s)
-
S_{\mathrm{fake}}(x_t,t,c)
\right)^\top
\frac{\partial \hat{x}_0}{\partial \theta}
\right].
\end{equation}
The teacher score uses classifier-free guidance,
\begin{equation}
\label{eq:cfg}
S^{\mathrm{CFG}}_{\mathrm{real}}(x_t,t,c;s)
=
S_{\mathrm{real}}(x_t,t,\emptyset)
+
s\left(
S_{\mathrm{real}}(x_t,t,c)-S_{\mathrm{real}}(x_t,t,\emptyset)
\right),
\end{equation}
where \(s\) is the guidance scale and \(c\) denotes the lip-sync conditioning inputs. We call \(s=1.0\) no-CFG and \(s>1\) CFG-guided sampling. Self Forcing samples the causal context \(\hat{x}^{<i}_0\) from the student’s own rollout rather than ground truth, reducing train-test exposure mismatch.

\section{Method}
\label{sec:method}

\subsection{Overview}
\label{sec:method_overview}
\modelname is a lip-sync-specific distillation framework that distills a high-fidelity bidirectional video-diffusion teacher (our lip-sync finetune of OmniAvatar~\cite{gan2025omniavatar},
OmniAvatar-LS; \supp{sup:teacher_model}) into a two-step causal student that accelerates streaming
lip synchronization while preserving reference fidelity.
Training proceeds in two stages.
We first pretrain the causal student via Diffusion Forcing (DF)~\cite{chen2024diffusion} on real data, with each chunk independently noised at a sampled timestep and supervised by the rectified flow matching objective (Sec.~\ref{sec:preliminaries}, \supp{sup:hyperparams}). 
This gives the causal student a clean conditional initialization for the subsequent distillation stage.
We then distill the pretrained student from the 14B teacher using Self Forcing DMD (Fig.~\ref{fig:architecture}) with three trajectory-analysis-derived modifications (Sec.~\ref{sec:method_analysis}): \emph{Sync-Window DMD} (SW-DMD; Sec.~\ref{sec:method_swdmd}), a two-step inference schedule (Sec.~\ref{sec:method_inference}), and a SyncNet-based reward (Sec.~\ref{sec:method_reward}). 

\begin{figure}[t]
    \centering
    \includegraphics[width=\linewidth]{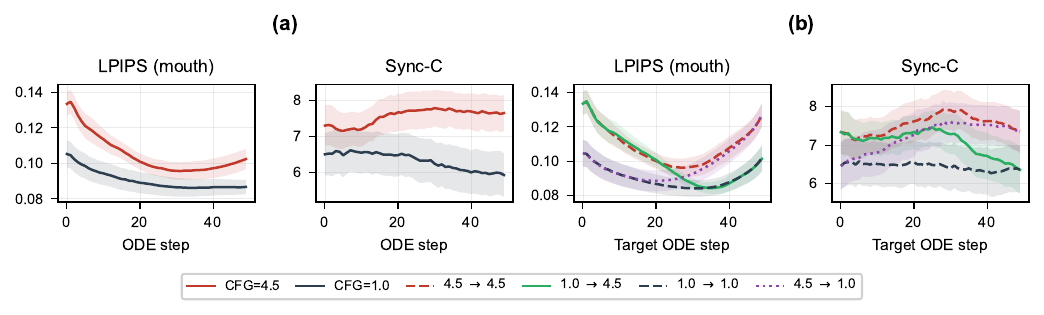}
    \caption{\textbf{Trajectory analysis of the 14B teacher.}  Bands are $\pm 1$~SE.
    \textbf{(a) CFG fidelity--sync tradeoff:} CFG ($s{=}4.5$, red) improves Sync-C but worsens reference fidelity (LPIPS), while no-CFG ($s{=}1.0$, navy) shows the opposite trend.
    \textbf{(b) Euler-step $2{\times}2$ factorial} over schedules $(s_0, s_1)$, plotted against the second-step landing $j_1$: mixed schedules recover most of the sync gap of the CFG-guided ceiling at landings near step~30.
    Full 4-metric versions in \supp{sup:trajectory_full}.}
    \label{fig:analysis_combined}
\end{figure}

\begin{figure}[htbp]
  \centering
  \includegraphics[width=\linewidth]{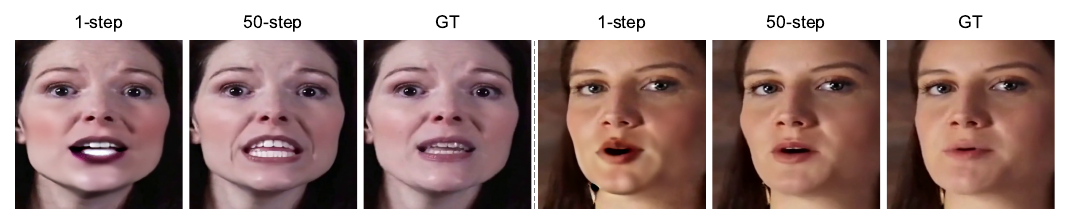}
  \caption{\textbf{Why few-step distillation needs trajectory-level care.}
  Two HDTF~\cite{zhang2021flow} samples, each showing the 1-step prediction from pure noise, 50-step ODE final output, and ground truth, respectively.
  Even a one-step prediction preserves coarse facial structure and approximate mouth timing, but it loses the fine articulation and audio-visual synchronization recovered by the full 50-step teacher.
  \modelname{} compresses this gap with a two-step student via the trajectory analysis of Sec.~\ref{sec:method_analysis}.}
  \label{fig:motivation}
\end{figure}

\subsection{Bidirectional teacher trajectory analysis}
\label{sec:method_analysis}
To identify and exploit the inherent characteristics of bidirectional lip synchronization models, we run a trajectory analysis on our 14B OmniAvatar-based teacher~\cite{gan2025omniavatar}.
On $n{=}10$ Hallo3~\cite{cui2025hallo3highlydynamicrealistic} clips held out from training, we save the per-step prediction $\hat{x}_0$ and evaluate reference fidelity (LPIPS~\cite{zhang2018unreasonable} on the mouth region) and audio-visual sync (SyncNet Sync-C~\cite{chung2016out}); setup details and additional results are in \supp{sup:analysis_setup_details}.
We analyze both the CFG-guided and no-CFG trajectories, as visualized in Fig.~\ref{fig:analysis_combined}(a).
Through this analysis we identify two characteristics of inpainting-based lip synchronization models.
First, lip-sync models exhibit a \emph{CFG fidelity--sync tradeoff}.
\label{sec:method_analysis_tradeoff}
Applying CFG improves audio-visual sync at the cost of reference fidelity, and vice versa for no-CFG inference.
Thus, no fixed CFG scale among those tested optimizes both metrics.
Second, because lip-sync tasks provide strong conditioning, even a one-step prediction preserves coarse facial structure and approximate mouth timing, though it lacks the fine articulation and audio-visual synchronization of the full 50-step teacher (Fig.~\ref{fig:motivation}).

\label{sec:method_analysis_factorial}
\begin{wrapfigure}{r}{0.40\linewidth}
    \centering
    \vspace{-1.4em}
    \includegraphics[width=\linewidth]{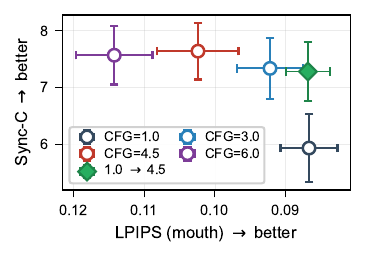}
    \vspace{-1.8em}
    \caption{\textbf{Fixed-CFG endpoints vs. diagnostic operating point} (green diamond,  at ODE step \(j=30\)).
    $n{=}10$, $\pm 1$~SE. SSIM and 4-metric in \supp{sup:trajectory_full}.}
    \label{fig:frontier}
    \vspace{-1em}
\end{wrapfigure}
\label{sec:method_analysis_frontier}
The strong one-step prediction suggests that the teacher does not require dense trajectory traversal for coarse lip timing and structure, but the remaining detail gap motivates asking where a second denoising step should land.
We therefore conduct an Euler-step analysis to simulate two-step teacher predictions.
For each schedule, we vary the second-step landing index $j_1$ using a single Euler step: from a shared near-pure-noise initial state $x_{\tau_0}$, take a CFG-guided or no-CFG velocity step to the candidate timestep $\tau_{j_1}$, then re-evaluate the teacher at $\tau_{j_1}$ with or without guidance.
Writing $s_0$ and $s_1$ for the guidance scales of the first and second teacher calls, we evaluate the four cells $(s_0, s_1) \in \{1.0, 4.5\}^2$ as in Fig.~\ref{fig:analysis_combined}(b).
Among them, the no-CFG$\to$CFG schedule at step~30 gives the best reference-sync compromise, occupying an operating point outside the fixed-CFG tradeoff (Fig.~\ref{fig:frontier}).
This direction is useful for distillation because reference degradation from CFG at the early step is difficult to undo, whereas the remaining sync gap can be reduced with explicit SyncNet supervision (Sec.~\ref{sec:method_reward}).
This diagnostic yields two design targets. Fig.~\ref{fig:analysis_combined}(a) identifies a broad sync-favoring band, roughly steps \(j\in[20,40]\), for training-time guidance, while Fig.~\ref{fig:analysis_combined}(b) selects step \(j=30\) as a representative landing for the two-step sampler. 
Crucially, this analysis characterizes the \emph{teacher's} guidance behavior: we transfer it
into the student through the distillation schedule (Sec.~\ref{sec:method_swdmd}) and the landing choice
(Sec.~\ref{sec:method_inference}), while the deployed student itself runs CFG-free.

\subsection{Sync-Window DMD}
\label{sec:method_swdmd}
\begin{figure}[t]
  \centering
  \includegraphics[width=\textwidth]{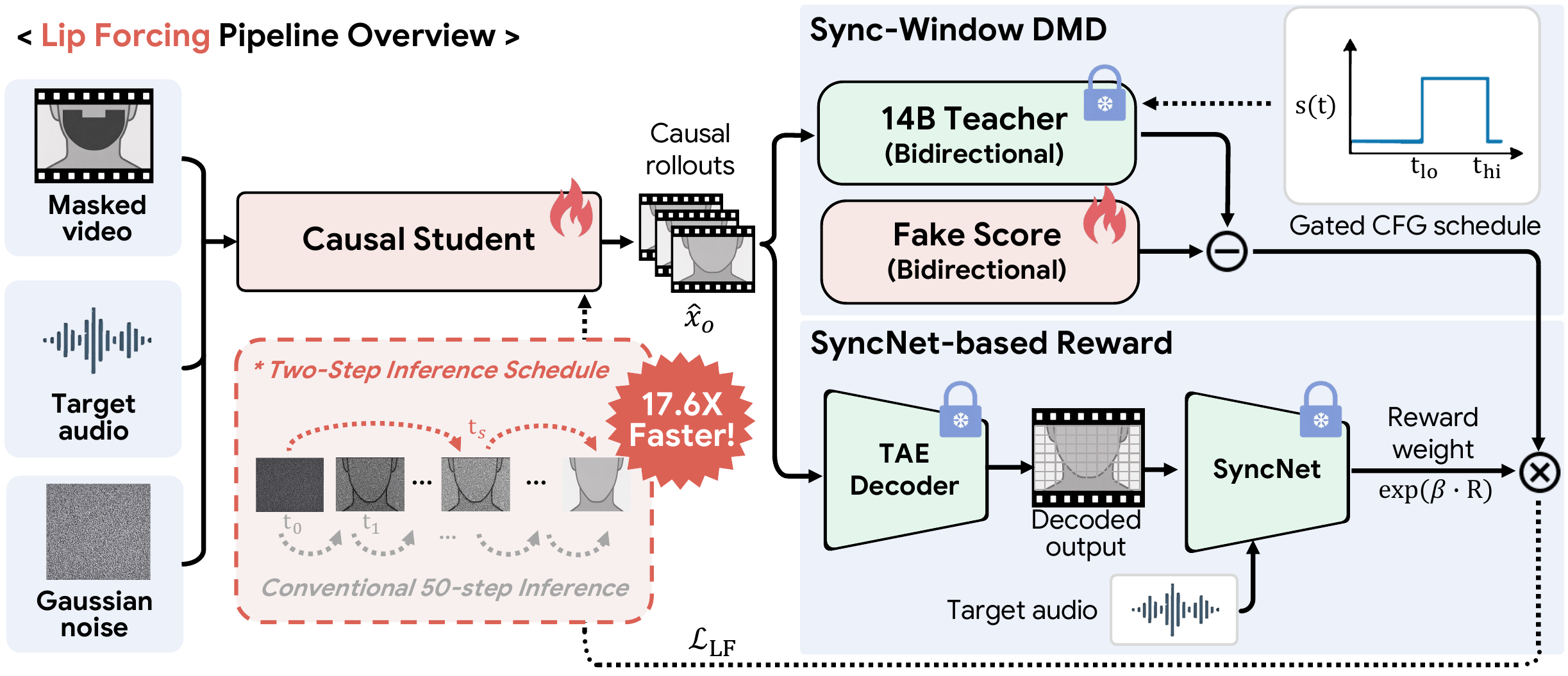}
  \vspace{-10pt}
\caption{\textbf{Architecture of \modelname.}
The causal student denoises Gaussian noise with lip-sync conditions, producing a chunk-wise causal rollout via the two-step schedule (Sec.~\ref{sec:method_inference}).
The clean prediction $\hat{x}_0$ is supervised by the DMD~\cite{yin2024one, yin2024improved} gradient (Eq.~\ref{eq:dmd_update}) between a frozen 14B teacher and a trainable fake-score critic, with the teacher's CFG gated by the windowed schedule $s_{\mathrm{SW}}$ of Eq.~\ref{eq:swdmd}.
The same $\hat{x}_0$ is decoded by the frozen Tiny AutoEncoder (TAE)~\cite{BoerBohan2025TAEHV} and scored by frozen SyncNet~\cite{chung2016out} against the conditioning audio to form the reward weight $\exp(\beta R)$ on the generator gradient (Eq.~\ref{eq:reward}).
}

  \label{fig:architecture}
\end{figure}

Standard DMD uses a fixed teacher CFG scale at every re-noising timestep. Motivated by Sec.~\ref{sec:method_analysis_factorial}, we instead use a timestep-gated teacher: no-CFG predictions preserve mouth-region fidelity, while CFG-guided predictions improve Sync-C over the mid-trajectory band where guidance is most useful for lip articulation, consistent with prior findings that middle
timesteps drive lip shape~\cite{he2025inpainting, peng2025omnisyncuniversallipsynchronization}.
Fig.~\ref{fig:architecture} summarizes our DMD pipeline.
We define \emph{Sync-Window DMD} (SW-DMD) by replacing the constant scale \(s\) in Eq.~\ref{eq:cfg} with a sync-window guidance schedule.
For a sampled DMD re-noising timestep \(t\) (the continuous timestep of Sec.~\ref{sec:preliminaries}),
let \(j(t)\) denote its corresponding index on the teacher's 50-step ODE grid. We set

\begin{equation}
\label{eq:swdmd}
s_{\mathrm{SW}}(j)=
\begin{cases}
4.5, & 20 \le j \le 40,\\
1.0, & \text{otherwise}.
\end{cases}
\end{equation}
The window \(20\le j\le40\) is a \emph{training guidance window}: DMD samples re-noising timesteps \(t \sim q(t)\), so the schedule covers the broad sync-improving band in Fig.~\ref{fig:analysis_combined}(a).
In contrast, inference uses a single representative second-step landing from the Euler-step plateau in Fig.~\ref{fig:analysis_combined}(b).
The timestep distribution \(q(t)\) and fake-score training are unchanged; only the teacher score uses \(s_{\mathrm{SW}}\).

\subsection{Two-step inference schedule}
\label{sec:method_inference}
At inference, the student uses two denoising model calls per chunk (plus one additional pass to cache the clean latent's KVs; \supp{sup:streaming_details}) and no inference-time CFG, at ODE indices \(J_{LF}=(0,30)\).
The first call denoises near-pure noise; the second call is the analysis-derived landing, after which Eq.~\ref{eq:x0_from_v} projects to the clean prediction \(\hat{x}_0\), which is then decoded and streamed.
The choice of the landing index is grounded in Sec.~\ref{sec:method_analysis}: at ODE step
\(j=30\), the diagnostic probe lands on the reference-leaning side of the joint reference-sync
optimum (Fig.~\ref{fig:frontier}), a deliberate choice to prioritize fidelity, with the residual sync gap reduced by the reward in Sec.~\ref{sec:method_reward}.
An earlier step would improve sync accuracy, whereas a later step would improve fidelity, exhibiting the tradeoff previously identified. 

\subsection{SyncNet-based reward}
\label{sec:method_reward}
Because the windowed schedule (SW-DMD) leaves the earlier steps at no-CFG, it retains a residual sync gap relative to the CFG-guided ceiling at training time, so we reduce the gap with an explicit sync reward adopted from Re-DMD~\cite{lu2025reward}.

We replace Re-DMD's video-dynamics reward with the SyncNet~\cite{chung2016out} confidence score $R(\cdot)$ between the conditioning audio $\mathbf{a}$ and the student's clean prediction $\hat{x}_0$, decoded by the lightweight Tiny AutoEncoder (TAE) decoder $D$~\cite{BoerBohan2025TAEHV}.
The reward is implemented in the DMD objective as a per-sample multiplicative weight on the generator gradient:
\begin{equation}
w(\hat{x}_0) = \exp\!\big(\beta \cdot R(D(\hat{x}_0),\, \mathbf{a})\big),
\qquad
\nabla_\theta \mathcal{L}_{\text{LF}}
\,\propto\,
w(\hat{x}_0) \cdot \nabla_\theta \mathcal{L}_{\text{DMD}},
\label{eq:reward}
\end{equation}
with $\beta = 2$ controlling reward strength.
$w(\hat{x}_0)$ is treated as a forward-only scalar: gradients flow through $\nabla_\theta \mathcal{L}_{\text{DMD}}$ only, and not through the reward function, the SyncNet model, or the TAE decoder.
Algorithm~\ref{alg:lipforcing} (\supp{sup:algorithm_extended}) gives the full \modelname training iteration.

\section{Experiments}
\label{sec:experiments}

\subsection{Experimental settings}
\label{sec:exp_settings}

\paragrapht{Implementation details.}
We instantiate \modelname at two student scales, both initialized from pretrained OmniAvatar~\cite{gan2025omniavatar} 1.3B and 14B backbones further finetuned for lip synchronization (OmniAvatar-LS; \supp{sup:teacher_finetune}).
The teacher signal in both instantiations is the 14B OmniAvatar-LS teacher; each student is paired with a learned fake-score critic sized to match the student.
Stage~2 (Self Forcing DMD distillation with our recipe) applies SW-DMD with CFG scale $4.5$ inside the windowed band and the SyncNet reward at strength $\beta{=}2$; at inference, the second step lands at $j{=}30$.
All training and inference use a fixed resolution of $512{\times}512$ and 81-frame sequences.
To optimize streaming performance, we use the Tiny AutoEncoder (TAE)~\cite{BoerBohan2025TAEHV} for the decoder and apply \texttt{torch.compile}.
FPS and TTFF are measured on a single NVIDIA H100 GPU.
The measurement methodology and a throughput--quality Pareto comparison against all baselines (Fig.~\ref{fig:pareto_all}) are provided in \supp{sup:compute_methodology}, with additional training and inference details in \supp{sup:hyperparams}.

\paragrapht{Datasets.}
We train our model on a mixture of VoxCeleb2~\citep{Chung_2018}, Hallo3~\citep{cui2025hallo3highlydynamicrealistic}, and HDTF~\citep{zhang2021flow}.
VoxCeleb2 provides diverse large-scale in-the-wild audio-visual pairs, which improves robustness and generalization.
HDTF and Hallo3 add higher-resolution facial videos and cleaner audio, offering richer facial details and more stable identity cues.
For evaluation, we use the HDTF test set of 33 clips.
Results on additional benchmarks~\cite{cui2025hallo3highlydynamicrealistic,chen2025talkvidlargescalediversifieddataset} and details on preprocessing are provided in \supp{sup:dataset_details}.

\paragrapht{Evaluation metrics.}
For visual quality, we measure FID~\citep{heusel2018ganstrainedtimescaleupdate} and SSIM~\citep{wang2004image}.
Additionally, we adopt FVD~\citep{unterthiner2019accurategenerativemodelsvideo} to measure temporal consistency of the generated video.
For identity preservation, we report CSIM, the cosine similarity between ArcFace~\citep{deng2019arcface} embeddings of generated and reference frames.
We assess lip-sync quality using the lip-sync confidence Sync-C and error distance Sync-D, computed with a pretrained expert~\citep{chung2016out}.
We also measure time-to-first-frame latency and throughput (FPS) of \modelname against the baselines to compare streaming performance during inference.

\subsection{Main comparison}
\label{sec:exp_main}

\begin{table}[t]
\centering
\small
\setlength{\tabcolsep}{3pt}
\caption{\textbf{Main comparison on HDTF~\cite{zhang2021flow}.} Quality, sync, and identity metrics across baselines and \modelname at two scales; TTFF in milliseconds. Best values \textbf{bold}; second-best \underline{underlined}.}
\label{tab:main_comparison}
\begin{tabular}{l c rr | ccc rr c}
\toprule
Method & Steps & FPS $\uparrow$ & TTFF $\downarrow$ & Sync-C $\uparrow$ & Sync-D $\downarrow$ & CSIM $\uparrow$ & FID $\downarrow$ & FVD $\downarrow$ & SSIM $\uparrow$ \\
\midrule
Ground truth                            & -- & -- & -- & 7.95 & 6.92 & --    & --    & --     & --    \\
\midrule
Wav2Lip~\cite{Prajwal_2020}                                                 & -- & \textbf{479.60} & \textbf{0.17} & \underline{8.56} & 6.70 & 0.946 & 24.15 & 384.82 & 0.911 \\
VideoReTalking~\cite{cheng2022videoretalkingaudiobasedlipsynchronization}    & -- & 2.67 & 3.76 & 8.22 & 6.70 & 0.910 & 24.59 & 306.63 & 0.883 \\
MuseTalk~\cite{zhang2025musetalkrealtimehighfidelityvideo}                   & 1 & 23.07 & 2.72 & 7.94 & 6.95 & \underline{0.957} &  9.68 & 127.44 & \underline{0.943} \\
\midrule
Diff2Lip~\cite{mukhopadhyay2023diff2lipaudioconditioneddiffusion}            & 25 & 15.47 & 5.04 & 8.35 & \underline{6.32} & 0.943  & 20.32 & 285.69 & 0.907  \\
LatentSync~\cite{li2024latentsync}                                     & 20& 3.23 & 6.29 & 8.10 & 6.51 & \textbf{0.967} &  6.90 & \underline{117.91} & \textbf{0.950} \\
X-Dub~\cite{he2025inpainting}                                                               & 30 & 0.91 & 163.64 & 7.58   & 7.66   & 0.898    & 14.76   & 183.99   & 0.831   \\
\midrule
OmniAvatar-LS (1.3B)~\cite{gan2025omniavatar}               & 50 & 1.79 & 45.36 & 8.04 & 6.99 & 0.927 &  8.06 & 143.75 & 0.904 \\
OmniAvatar-LS (14B)~\cite{gan2025omniavatar}                & 50 & 0.38 & 213.72 & \textbf{8.98} & \textbf{6.11} & 0.934 &  \textbf{6.71} & 133.87 & 0.911 \\
\midrule
Self Forcing (1.3B)~\cite{huang2025self}                           & 4  & 27.48 & 0.38 &  7.12  & 7.80   & 0.939    & 7.51    & 124.78     & 0.915    \\
\modelname (1.3B, Ours)                       & 2  & \underline{31.58} & \underline{0.32} & 6.88 & 7.93 & 0.943 &  \underline{6.76} & 118.86 & 0.919 \\
\modelname (14B, Ours)                       & 2  & 15.11 & 0.54 & 7.59 & 7.23 & 0.949 &  7.01 & \textbf{107.88} & 0.938 \\
\bottomrule
\end{tabular}

\end{table}

We compare \modelname against prior lip-sync methods on the HDTF~\cite{zhang2021flow} test set (Tab.~\ref{tab:main_comparison}), with a vanilla Self Forcing DMD 1.3B baseline included to isolate the recipe from generic distillation gains.

\modelname (1.3B) is the fastest diffusion method in the table, exceeding the $25$~FPS playback rate of the test videos. The Self Forcing baseline is the only other diffusion method to cross this threshold, while every multi-step diffusion baseline runs below the real-time threshold.
TTFF is sub-millisecond at both \modelname scales, an order of magnitude below the fastest multi-step diffusion baseline.
Against the same-init bidirectional models OmniAvatar-LS~\cite{gan2025omniavatar}, \modelname is $17.6\times$ and $39.8\times$ faster at 1.3B and 14B, respectively, and the 14B student is also $4.7\times$ faster than LatentSync~\cite{li2024latentsync}.

\begin{figure}[t]
  \centering
  \includegraphics[width=\linewidth,keepaspectratio]{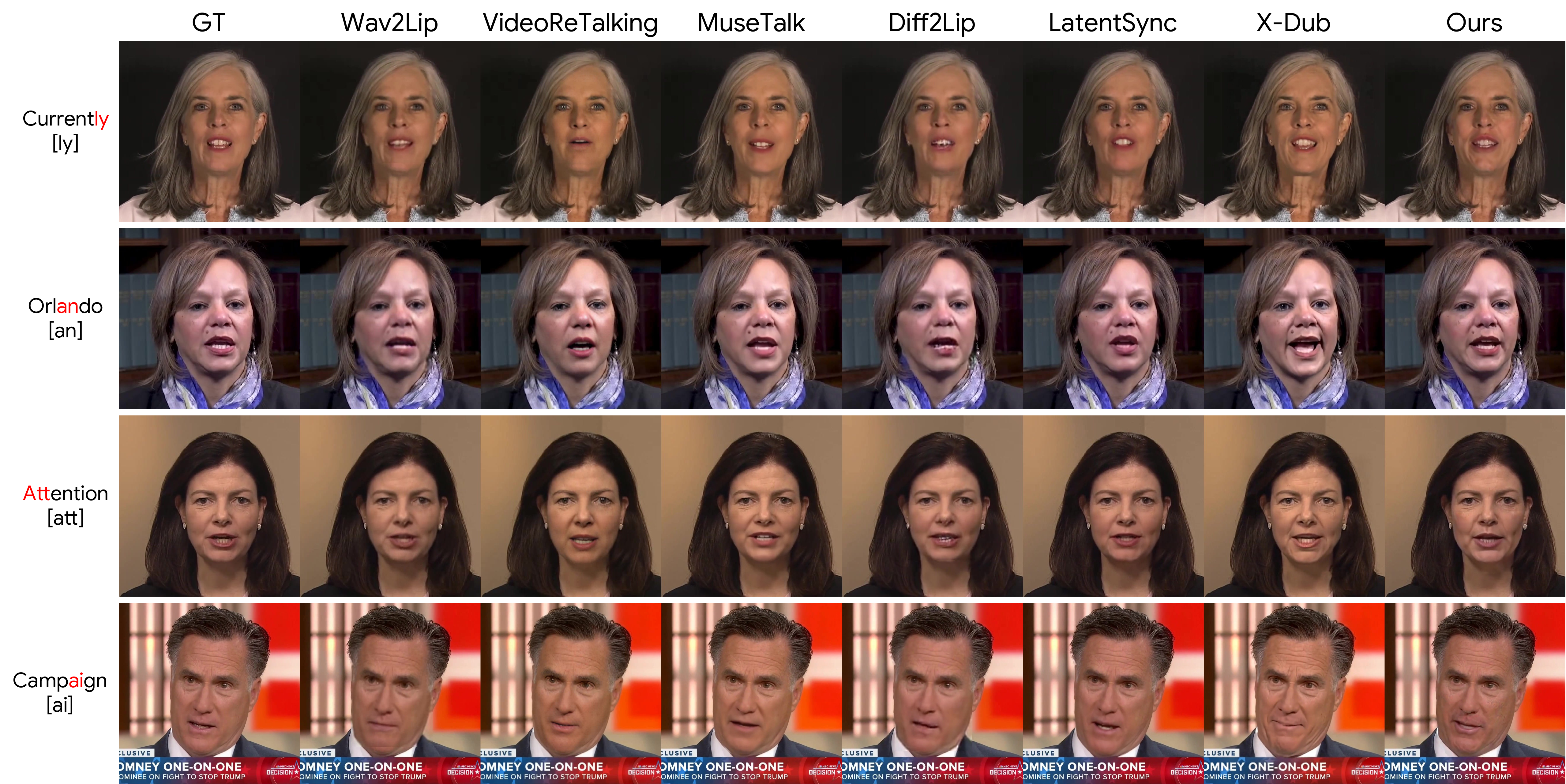}
  \caption{\textbf{Qualitative comparison on HDTF.} Each row shows the same source frame rendered by our method, six lip-sync baselines, and the ground truth (GT) at the moment of articulating the bracketed English phoneme. Best viewed zoomed in and in color.}
  \label{fig:qualitative}
\end{figure}

Qualitative comparisons against all baselines at matched phoneme-articulation moments are shown in Fig.~\ref{fig:qualitative}.

By design, \modelname operates on the reference-leaning side of the CFG fidelity--sync tradeoff (Sec.~\ref{sec:method_inference}): the analysis-derived landing at $j{=}30$ and the capped SyncNet reward deliberately exchange a portion of audio-visual sync for reference fidelity.
This is visible throughout Tab.~\ref{tab:main_comparison} (strong FID, FVD, and identity while trailing the strongest baselines on Sync-C), and the user study (Sec.~\ref{sec:exp_user_study}) indicates this gap reflects the SyncNet metric more than perceived synchronization.

Against single-pass baselines~\cite{Prajwal_2020,cheng2022videoretalkingaudiobasedlipsynchronization,zhang2025musetalkrealtimehighfidelityvideo}, \modelname trades a Sync-C deficit for substantially lower FID and FVD; the strongest single-pass peer MuseTalk leads on CSIM and SSIM.
Wav2Lip and VideoReTalking exceed the ground-truth Sync-C, suggesting these objectives over-fit the SyncNet expert at the cost of perceptual realism, a tendency the user study (Sec.~\ref{sec:exp_user_study}) probes directly.

Against multi-step diffusion baselines~\cite{mukhopadhyay2023diff2lipaudioconditioneddiffusion,li2024latentsync,he2025inpainting}, \modelname is on par with LatentSync on FVD at 1.3B and posts the best FVD overall at 14B, dominating Diff2Lip and X-Dub on every fidelity metric and strictly improving over the same-init OmniAvatar-LS parent on FVD, SSIM, and CSIM.
Against the Self Forcing 1.3B baseline~\cite{huang2025self}, which lacks SW-DMD, the analysis-derived two-step landing, and the SyncNet reward, \modelname improves every fidelity and identity metric with a minor Sync regression at half the step count, consistent with the reference-leaning operating point above.
At 14B the Sync-C gap closes further toward the ground-truth value, indicating that recipe gains compound with scale.
Long-form rollouts and cross-identity audio evaluations are reported in \supp{sup:long_video_extended} and \supp{sup:cross_audio_extended}.

\subsection{Ablations}
\label{sec:exp_ablations}

In the following, we ablate the main components (Sec.~\ref{sec:method_swdmd}, 
Sec.~\ref{sec:method_reward}) and key design choices: CFG schedule shape, step count, and second-step landing.
We focus on FVD as the primary fidelity axis, since SSIM is comparable across configurations.

\paragrapht{Components.}
We ablate the CFG schedule and SyncNet reward at the two-step inference schedule (Tab.~\ref{tab:components}) to isolate each component's contribution.
Switching from static CFG to the windowed schedule~(Eq.~\ref{eq:swdmd}) substantially improves FVD at the cost of a small Sync regression, while adding the SyncNet reward~(Eq.~\ref{eq:reward}) consistently improves Sync-C in both CFG settings.
The full \modelname recipe (windowed $+\,R$) therefore balances both axes: SW-DMD recovers the fidelity that static CFG sacrifices for sync, and the explicit reward narrows the residual sync gap that windowing introduces.

\paragrapht{CFG schedule shape.}
We compare the windowed schedule against fixed-CFG endpoints and a reverse-direction window at the two-step inference schedule, with the SyncNet reward disabled (Tab.~\ref{tab:cfg_schedule}).
The all-CFG and no-CFG endpoints bracket the fidelity--sync tradeoff identified in Sec.~\ref{sec:method_analysis_frontier}: all-CFG drives the strongest sync at the worst FVD, while no-CFG inverts both directions.
Windowing CFG only inside the analysis-derived sync-favoring band sacrifices a small amount of all-CFG's sync to deliver the best FVD in the table; the reverse-direction window confirms the placement by exhibiting the opposite tradeoff, picking up a sliver of sync over windowed at the cost of fidelity.

\paragrapht{Step count.}
We compare \modelname's 2-step inference at $J_{LF}{=}(0,30)$ against 1-step, a uniform 2-step at $j_1{=}25$, and a 4-step variant ($J_{LF}{=}(0,13,25,37)$) of our recipe (Tab.~\ref{tab:steps}).
Most metrics are comparable, and the step count primarily moves FVD.
The 4-step variant achieves the best FVD and serves as the recipe's full-trajectory reference, while the 1-step setting has the worst FVD.
\modelname's 2-step at $j_1{=}30$ closes most of the 1-step vs. 4-step FVD gap at half the 4-step inference cost; comparison with the uniform $j_1{=}25$ landing is further analyzed in Tab.~\ref{tab:second_step}.

\paragrapht{Second-step landing.}
\label{sec:exp_ablations_t}
We sweep the second-step landing $j_1$ across ours and those used by Self Forcing~\cite{huang2025self}.
Following the analysis in Fig.~\ref{fig:analysis_combined}, $j_1{=}25$ gives the best sync performance at a slight FVD cost, and $j_1{=}37$ inverts the tradeoff (best FVD, lowest Sync-C); SSIM remains comparable across all four landings.
Our setting of $j_1{=}30$ balances the two, and the early $j_1{=}13$ landing is suboptimal on both axes.
The second-step landing therefore offers a direct knob for trading fidelity against sync.

\begin{table}[t]
\centering
\footnotesize
\setlength{\tabcolsep}{3pt}
\begin{minipage}[t]{0.48\linewidth}
\centering
\caption{\textbf{Component ablation.} CFG schedule (\emph{static}: CFG$=$4.5; \emph{windowed}: Eq.~\ref{eq:swdmd}) and SyncNet reward $R$ at two-step inference. Bold row: full \modelname.}
\label{tab:components}
\vspace{4pt}
\begin{tabular}{l ccrc}
\toprule
Config. & Sync-C $\uparrow$ & Sync-D $\downarrow$ & FVD $\downarrow$ & SSIM $\uparrow$ \\
\midrule
static                  & 7.13 & 7.85 & 138.32 & 0.916 \\
static $+ R$            & \textbf{7.24} &\textbf{7.76} & 135.94 & 0.917 \\
windowed                & 6.81 & 7.85 & 119.88 & \textbf{0.920} \\
\textbf{windowed $\mathbf{+\,R}$} & 6.88 & 7.93 & \textbf{118.86} & 0.919 \\
\bottomrule
\end{tabular}
\end{minipage}

\hfill
\begin{minipage}[t]{0.48\linewidth}
\centering
\caption{\textbf{CFG schedule ablation.} Two-step inference, no reward. \emph{all-CFG}: CFG$=$4.5 everywhere; \emph{no-CFG}: CFG$=$1.0 everywhere; \emph{reverse}: CFG outside the window, no-CFG inside.}
\label{tab:cfg_schedule}
\vspace{4pt}
\begin{tabular}{l ccrc}
\toprule
Schedule & Sync-C $\uparrow$ & Sync-D $\downarrow$ & FVD $\downarrow$ & SSIM $\uparrow$ \\
\midrule
all-CFG             & \textbf{7.13} & 7.85 & 138.32 & 0.916 \\
no-CFG              & 6.14 & 8.39 & 120.85 & \textbf{0.921} \\
\textbf{windowed}   & 6.81 & 7.85 & \textbf{119.88} & 0.920 \\
reverse             & 6.98 & \textbf{7.81} & 126.62 & 0.917 \\
\bottomrule
\end{tabular}
\end{minipage}

\vspace{1em}

\begin{minipage}[t]{0.48\linewidth}
\centering
\caption{\textbf{Step count ablation.} Windowed CFG schedule, no reward. Step indices: 1-step; uniform 2-step; ours 2-step; ours 4-step.}
\label{tab:steps}
\vspace{4pt}
\begin{tabular}{l ccrc}
\toprule
\# of Steps & Sync-C $\uparrow$ & Sync-D $\downarrow$ & FVD $\downarrow$ & SSIM $\uparrow$ \\
\midrule
1                          & 6.81 & 7.92 & 131.50 & \textbf{0.926} \\
2 (uniform, $j_1{=}25$)    & \textbf{6.95} & \textbf{7.85} & 124.57 & \textbf{0.926} \\
\textbf{2 (Ours, $\mathbf{j_1{=}30}$)} & 6.81 & \textbf{7.85} & 119.88 & 0.920 \\
4                    & 6.81 & 8.01 & \textbf{117.80} & 0.923 \\
\bottomrule
\end{tabular}
\end{minipage}

\hfill
\begin{minipage}[t]{0.48\linewidth}
\centering
\caption{\textbf{Second-step landing ablation.} Two-step inference at $J_{LF}{=}(0, j_1)$, windowed CFG schedule, no reward.}
\label{tab:second_step}
\vspace{4pt}
\begin{tabular}{l ccrc}
\toprule
$j_1$ & Sync-C $\uparrow$ & Sync-D $\downarrow$ & FVD $\downarrow$ & SSIM $\uparrow$ \\
\midrule
13                  & 6.79 & 7.92 & 135.22 & \textbf{0.927} \\
25                  & \textbf{6.95} & \textbf{7.85} & 124.57 & 0.926 \\
\textbf{30}  & 6.81 & \textbf{7.85} & 119.88 & 0.920 \\
37                  & 6.73 & 7.87 & \textbf{114.78} & 0.920 \\
\bottomrule
\end{tabular}
\end{minipage}

\end{table}

\subsection{User study}
\label{sec:exp_user_study}

\begin{wraptable}{r}{0.54\linewidth}
\vspace{-\intextsep}
\centering
\small
  \caption{\textbf{User study.} Mean Opinion Score (1--5 Likert, higher is better) on four axes: video--audio synchronization (Sync), video quality (Qual.), identity preservation (ID), and naturalness (Nat.). Best per column \textbf{bold}; second-best \underline{underlined}.}
  \label{tab:user_study}
  \setlength{\tabcolsep}{4pt}
  \begin{tabular}{l cccc}
  \toprule
  Method & Sync $\uparrow$ & Qual. $\uparrow$ & ID $\uparrow$ & Nat. $\uparrow$ \\
  \midrule
  Wav2Lip~\cite{Prajwal_2020}                                                 & 3.43 & 2.60 & 3.32 & 2.75 \\
  VideoReTalking~\cite{cheng2022videoretalkingaudiobasedlipsynchronization}    & 3.49 & 3.00 & 3.43 & 3.21 \\
  MuseTalk~\cite{zhang2025musetalkrealtimehighfidelityvideo}                   & 3.47 & 3.34 & 3.56 & 3.16 \\
  Diff2Lip~\cite{mukhopadhyay2023diff2lipaudioconditioneddiffusion}            & 3.15 & 2.25 & 3.12 & 2.47 \\
  LatentSync~\cite{li2024latentsync}                                           & 3.96 & 3.54 & 3.82 & 3.53 \\
  X-Dub~\cite{he2025inpainting}                                                & \textbf{4.40} & \underline{4.13} & \underline{4.25} & \underline{3.97} \\
  \modelname{} (14B)                                                          & \underline{4.38} & \textbf{4.33} & \textbf{4.46} & \textbf{4.32} \\
  \bottomrule
  \end{tabular}
\end{wraptable}

We run a Mean Opinion Score (MOS) user study comparing \modelname against six lip-sync baselines on a 30-clip pool drawn from HDTF~\cite{zhang2021flow} and TalkVid~\cite{chen2025talkvidlargescalediversifieddataset}.
Each anonymized output is scored on four 5-point Likert items: video-audio synchronization, video quality, identity preservation, and naturalness (Tab.~\ref{tab:user_study}). 
The full protocol is detailed in \supp{sup:user_study_protocol}.
We find that our quality, ID, and naturalness MOS scores exceed those of all baselines, in line with the fidelity metrics. Our model scores on par with the highest-performing baseline in sync, which we attribute to the high overall quality of our model, and argue that the dip in sync metrics has minimal negative effects on user experience.

\section{Conclusion}
\label{sec:conclusion}

We presented \modelname, a streaming lip-synchronization framework that distills a bidirectional video-diffusion teacher into a two-step causal autoregressive student through a trajectory-analysis-derived framework: Sync-Window DMD, an analysis-derived two-step inference schedule, and a SyncNet-based reward.
Distilled from a single 14B OmniAvatar-based teacher, \modelname (1.3B) enables real-time streaming with sub-millisecond time-to-first-frame, and \modelname (14B) posts the best FVD in our comparison and approaches the ground-truth Sync-C value, while remaining $4.7\times$ faster than LatentSync at comparable reference fidelity.

These speed and quality numbers bring streaming lip synchronization within reach of latency-sensitive applications such as live translation, virtual avatars, and interactive agents.
More broadly, the trajectory-aware diagnostic that drives the recipe is a general procedure for adapting a conditional-diffusion lip-sync teacher to few-step distillation: it identifies the teacher's guidance structure where present and parameterizes a corresponding recipe, transferring as a methodology rather than as fixed cutoffs.

{
\small

\bibliographystyle{plainnat}
\bibliography{neurips_2026}

}

\clearpage
\appendix
\section*{\Large Appendix}

\section{Index of supplementary material}
\label{sup:index}

This appendix is organized into seven sections.
Section~\ref{sup:teacher_model} describes the teacher model;
Sec.~\ref{sup:analysis_extended} extends the trajectory analysis of Sec.~\ref{sec:method_analysis};
Sec.~\ref{sup:method_extended} provides reproducibility detail for Sec.~\ref{sec:method};
Sec.~\ref{sup:experiments_extended} reports auxiliary evaluations and methodology supplementing Sec.~\ref{sec:experiments};
Sec.~\ref{sup:additional_qual} collects additional qualitative results;
and Secs.~\ref{sup:limitations} and~\ref{sup:broader_impact} discuss limitations and societal impact.

\paragrapht{Section~\ref{sup:teacher_model} (Teacher model).}
\begin{itemize}\itemsep0pt
    \item \S\ref{sup:omniavatar_overview}: original OmniAvatar architecture and audio conditioning.
    \item \S\ref{sup:teacher_finetune}: lip-sync finetuning (OmniAvatar-LS), including the lip-region mask convention.
    \item \S\ref{sup:teacher_data}: training data and preprocessing pipeline (shared with the distilled student).
    \item \S\ref{sup:teacher_training}: teacher training details.
\end{itemize}

\paragrapht{Section~\ref{sup:analysis_extended} (Analysis, extended).}
\begin{itemize}\itemsep0pt
    \item \S\ref{sup:analysis_setup_details}: trajectory analysis setup details (the $j \leftrightarrow \tau_j$ mapping, clip selection, metric implementations).
    \item \S\ref{sup:trajectory_full}: full 4-metric versions of main Fig.~\ref{fig:analysis_combined} and Fig.~\ref{fig:frontier}.
    \item \S\ref{sup:audio_only_drop}: audio-only CFG drop variant cross-checks.
    \item \S\ref{sup:plateau}: trajectory plateau details and paired $t$-test.
\end{itemize}

\paragrapht{Section~\ref{sup:method_extended} (Method, extended).}
\begin{itemize}\itemsep0pt
    \item \S\ref{sup:hyperparams}: hyperparameters and training details.
    \item \S\ref{sup:syncnet_reward_details}: SyncNet reward implementation details.
    \item \S\ref{sup:streaming_details}: streaming rollout details.
    \item \S\ref{sup:algorithm_extended}: full algorithm pseudocode.
\end{itemize}

\paragrapht{Section~\ref{sup:experiments_extended} (Experiments, extended).}
\begin{itemize}\itemsep0pt
    \item \S\ref{sup:compute_methodology}: compute and efficiency measurement methodology.
    \item \S\ref{sup:user_study_protocol}: user study protocol.
    \item \S\ref{sup:dataset_details}: datasets and additional benchmarks (HDTF, Hallo3, TalkVid).
    \item \S\ref{sup:long_video_extended}: long-video evaluation.
    \item \S\ref{sup:cross_audio_extended}: cross-identity audio evaluation.
\end{itemize}

\paragrapht{Section~\ref{sup:additional_qual} (Additional Qualitative Results).}
Additional side-by-side baseline comparisons on the Hallo3, HDTF, and TalkVid test sets.

\paragrapht{Section~\ref{sup:limitations} (Limitations) and Section~\ref{sup:broader_impact} (Broader Impact).}
Discussion of recipe scope, generalization, and societal implications.

\section{Teacher model}
\label{sup:teacher_model}

This appendix describes the teacher diffusion model that supervises our distilled student in Sec.~\ref{sec:method}.
We start from OmniAvatar~\cite{gan2025omniavatar}, an audio-driven portrait animation model, and finetune it for the inpainting-based lip-sync task that the rest of the paper targets.
Sec.~\ref{sup:omniavatar_overview} recaps the original OmniAvatar architecture; Sec.~\ref{sup:teacher_finetune} details our finetuning recipe (referred to as OmniAvatar-LS in the main paper); Secs.~\ref{sup:teacher_data} and~\ref{sup:teacher_training} describe the training data pipeline shared with the distilled student and the teacher-side training hyperparameters, respectively.

\subsection{OmniAvatar overview}
\label{sup:omniavatar_overview}

OmniAvatar is an audio-driven portrait animation model in the image-to-video (I2V) family: given a reference image and a corresponding audio clip, it generates a short video of the subject speaking the audio.
The backbone is the Wan~2.1 video diffusion transformer~\cite{wan2025}, released at 1.3B and 14B parameters; OmniAvatar adopts both scales and leaves the transformer blocks unchanged.

Audio is injected through an Audio Pack module rather than the conventional audio cross-attention layers used in earlier portrait animation models.
The audio is first encoded by Wav2Vec~2.0~\cite{baevski2020wav2vec20frameworkselfsupervised}, then projected into the same latent dimensionality as the video tokens, and finally added directly to the noisy video latents at the input of the transformer.
On the visual side, the noisy target video latents are concatenated along the channel dimension with the reference image (encoded by the frozen Wan 3D VAE and broadcast across the temporal axis) and a binary mask whose temporal axis designates the first frame as fixed conditioning and all subsequent frames as to be inpainted; the resulting tensor is the visual input to the transformer.

The transformer is finetuned with LoRA~\cite{hu2022lora} adapters applied to the attention and FFN layers; all other parameters remain frozen.
For classifier-free guidance, OmniAvatar randomly drops the audio condition with $10\%$ probability during finetuning, while text-drop CFG is inherited from the Wan~2.1 backbone.
Standard inference applies both text and audio guidance jointly, with their unconditional branches dropped together at scale~$4.5$; because the two drops are introduced at different stages of training, the model also supports dropping audio independently of text, which our trajectory analysis exploits (Sec.~\ref{sec:method_analysis}).

We adopt OmniAvatar as our teacher because it combines three properties our distillation target needs: a strong pretrained Wan~2.1 video prior at the 14B scale, direct audio conditioning via the Audio Pack, and a guidance scheme rich enough to support the audio-only and audio+text drop variants used in our trajectory analysis (Sec.~\ref{sec:method_analysis}).

\subsection{Lip-sync finetuning (OmniAvatar-LS)}
\label{sup:teacher_finetune}

We adapt OmniAvatar from its original I2V portrait animation setting to the video-conditioned lip-sync task that the rest of the paper targets, keeping the diffusion transformer and the LoRA finetuning recipe of Sec.~\ref{sup:omniavatar_overview} unchanged.
The only modifications are in the input pipeline: \emph{what} is concatenated into the model, and \emph{how} the noise mask is interpreted.

\paragrapht{Inputs to the model.}
Where OmniAvatar concatenates only a static reference image and binary temporal mask with the noisy target video latents, our teacher concatenates five video-shaped tensors along the channel dimension after each is processed by the appropriate frozen encoder.
For an $81$-frame, $512{\times}512$ training clip, the Wan 3D VAE's $4{\times}$ temporal and $8{\times}$ spatial compression yields $21{\times}64{\times}64$ latents; the five concatenated inputs are listed below with their channel counts $C_i$:
\begin{itemize}\itemsep0pt
    \item \textbf{Input video latents} ($C_i{=}16$): the latents of the target video (Gaussian noise at inference), encoded by the Wan 3D VAE.
    \item \textbf{Mask} ($C_i{=}1$): the lip-region binary mask, downsampled from pixel resolution to the latent grid and broadcast across the temporal axis (described in detail below).
    \item \textbf{Reference frame} ($C_i{=}16$): a single frame randomly sampled from the source clip, encoded by the 3D VAE and broadcast across time, identical in role to the OmniAvatar reference image.
    \item \textbf{Reference masked video} ($C_i{=}16$): the input video with the lip region zeroed out, encoded by the 3D VAE; this provides ground-truth context for the unmasked area at every frame.
    \item \textbf{Reference frame sequence} ($C_i{=}16$): a short clip sampled from the same source video that does not overlap with the input window, encoded by the 3D VAE; it supplies additional identity and motion priors as in conventional video-conditioned lip-sync models~\cite{Prajwal_2020, li2024latentsync}.
\end{itemize}
The five tensors total $\sum_i C_i = 65$ channels per spatiotemporal latent voxel, so the visual input fed into the diffusion transformer has shape $[B, 65, 21, 64, 64]$.
The Audio Pack output is added to this concatenated tensor exactly as in OmniAvatar, so the audio conditioning pathway is otherwise unchanged.

\paragrapht{Mask interpretation.}
The original OmniAvatar mask designates the first frame as \emph{fixed} (clean conditioning) and all subsequent frames as \emph{to be inpainted} (noised), so generation flows temporally outwards from a single anchor frame.
We replace this temporal scheme with a spatial one driven by the lip-region mask: at every frame, pixels inside the mask are treated as \emph{to be inpainted} while pixels outside the mask are treated as \emph{fixed}.
The model therefore preserves the input video everywhere outside the lip region and only regenerates the masked area conditioned on audio, matching the inpainting formulation used by the distilled student in Sec.~\ref{sec:method}.

\paragrapht{Lip-region mask.}
We adopt the U-shaped lip-region mask convention from LatentSync~\cite{li2024latentsync}, covering the mouth, chin, and lower face along the jaw line; the geometry is shown in the \emph{Mask} block of the architecture overview.
The mask is resized to the same $512{\times}512$ pixel resolution as the input video and enters the model along two paths.
First, it is applied to the original RGB video frames in pixel space; the resulting masked video is then encoded by the frozen Wan 3D VAE to produce the \emph{reference masked video} latents listed above.
Second, the mask itself is downsampled to the $64{\times}64$ Wan latent grid and broadcast across the temporal axis to form the \emph{mask} channel of the input concatenation; the mask is never encoded by the VAE.
The geometry is fixed across all training and evaluation clips because the input frames are face-aligned to a canonical pose during preprocessing (Sec.~\ref{sup:teacher_data}), so no per-clip mask estimation is required at inference.

\subsection{Training data and preprocessing}
\label{sup:teacher_data}

Teacher finetuning (Sec.~\ref{sup:teacher_finetune}) and student distillation (Sec.~\ref{sec:method}) share an identical data processing pipeline; we describe it once here for both stages.

\paragrapht{Training datasets.}
We train on a mixture of three audio-visual datasets:
\begin{itemize}\itemsep0pt
    \item \textbf{VoxCeleb2}~\cite{Chung_2018}: a large-scale audio-visual speaker dataset collected from YouTube interview videos, with over 1M utterances from 6{,}000+ speakers spanning a wide range of ethnicities, accents, and recording conditions; we use a 50K-clip random subsample.
    \item \textbf{HDTF}~\cite{zhang2021flow}: $\sim$362 in-the-wild talking-face videos totaling 15.8\,hours at 720p/1080p, providing high visual quality and clean audio for stable identity cues.
    \item \textbf{Hallo3}~\cite{cui2025hallo3highlydynamicrealistic}: 70+ hours of talking-head videos plus 50+ wild-scene clips, contributing dynamic backgrounds and varied camera viewpoints.
\end{itemize}

\paragrapht{Preprocessing.}
We follow the LatentSync~\cite{li2024latentsync} pipeline.
Videos are resampled to 25\,fps and audio to 16\,kHz; scene detection segments each video at shot boundaries into 5--10\,s clips.
Faces are detected and aligned with InsightFace~\cite{deng2019arcface}: an affine transformation maps facial landmarks onto a canonical pose, after which each frame is resized to $512{\times}512$.

\paragrapht{Filtering.}
Two filters remove low-quality clips.
SyncNet~\cite{chung2016out} confidence is computed on each clip and clips below~$3$ are discarded; the audio-visual offset is adjusted to~$0$ for the remainder.
HyperIQA~\cite{su2020blindly} scores are then computed on the surviving clips and clips below~$40$ are removed.
The final pool contains approximately $30$K clips and is used identically by both training stages.

\paragrapht{Clip and reference sampling.}
Each training step draws an $81$-frame input window ($\sim 3.24$\,s at $25$\,fps) from a uniformly sampled clip in the filtered pool.
The reference frame is sampled uniformly at random from anywhere in the same source clip and broadcast across the temporal axis after VAE encoding, matching the OmniAvatar reference-image conditioning.
The reference frame sequence is a separate $81$-frame window sampled with a random start \emph{outside} the input window of the same source clip, providing non-redundant identity and motion priors.
For source clips shorter than $162$ frames where fully disjoint windows are not available, the reference sequence is permitted to overlap minimally with the input window.

\subsection{Teacher training details}
\label{sup:teacher_training}

We finetune OmniAvatar-LS at 1.3B and 14B parameter scales; both are initialized from the public OmniAvatar release weights at the corresponding scale and adapted to the input pipeline and lip-region mask of Sec.~\ref{sup:teacher_finetune}.
The 14B model serves as the distillation teacher (Sec.~\ref{sec:method}), while the 1.3B model provides the 1.3B student's initialization and a same-scale baseline (Tab.~\ref{tab:main_comparison}).

\paragrapht{Setup.}
Training uses Hugging Face Accelerate~\cite{accelerate} with PyTorch DDP at \texttt{bf16} mixed precision (no FSDP or DeepSpeed) on NVIDIA H200 GPUs.
The 14B teacher runs on 4 GPUs with per-device batch size $1$ and gradient accumulation $2$; the 1.3B teacher runs on 2 GPUs with per-device batch size $1$ and gradient accumulation $4$, both yielding an effective batch size of $8$.
Wall-clock time is approximately $1$ week for the 14B teacher and $3$ days for the 1.3B teacher; project-level compute totals are reported in Sec.~\ref{sup:hyperparams}.
We use AdamW~\cite{loshchilov2017decoupled} with PyTorch defaults ($\beta_1{=}0.9$, $\beta_2{=}0.999$, $\epsilon{=}10^{-8}$), weight decay $0.01$, a constant learning rate of $5{\times}10^{-5}$, and gradient norm clipped to~$1.0$.
LoRA~\cite{hu2022lora} adapters of rank $128$ and scale $\alpha{=}64$ are added to the attention and FFN layers following the OmniAvatar~\cite{gan2025omniavatar} convention; all other parameters remain frozen.

\paragrapht{Training objective.}
The primary loss is mean-squared error on the rectified-flow velocity prediction, with a mouth-region weight of $2.0$ applied inside the lip-region mask of Sec.~\ref{sup:teacher_finetune} so that the velocity supervision is biased toward the inpainted area.
Three auxiliary losses, computed on the $\hat{x}_0$ prediction obtained by decoding the predicted clean latent through the frozen Wan 3D VAE, supplement the velocity term: SyncNet~\cite{chung2016out} confidence (weight~$0.05$), LPIPS~\cite{zhang2018unreasonable} perceptual loss (weight~$0.15$), and the TREPA~\cite{li2024latentsync} temporal-representation alignment term (weight~$10.0$).
For classifier-free guidance, audio and text are independently dropped at $10\%$ probability per step, supporting both joint and audio-only drop modes at inference (Sec.~\ref{sup:omniavatar_overview}).

\section{Analysis (extended)}
\label{sup:analysis_extended}

\subsection{Trajectory analysis setup details}
\label{sup:analysis_setup_details}

This subsection provides implementation details for the trajectory analysis of Sec.~\ref{sec:method_analysis}.

\paragrapht{Teacher and clips.}
The trajectory analysis uses the 14B OmniAvatar-LS teacher (Sec.~\ref{sup:teacher_model}) on $n{=}10$ Hallo3~\cite{cui2025hallo3highlydynamicrealistic} clips held out from training.
Clips are processed at $512{\times}512$ resolution and 81 frames at $25$\,fps, matching the data pipeline of Sec.~\ref{sup:teacher_data}.
The same clip set, noise seed, and reference frame are used across all CFG variants so that per-step trajectories can be compared as paired samples.

\paragrapht{Shifted-ODE schedule.}
The teacher uses OmniAvatar's rectified-flow schedule with $N{=}50$ inference steps and a non-uniform timestep mapping that concentrates samples at high noise levels.
Uniform timesteps are
\begin{equation}
u_j = u_{\max} - j \cdot \frac{u_{\max} - u_{\min}}{N},
\qquad j = 0, 1, \ldots, N,
\label{eq:uniform_t}
\end{equation}
with $u_{\max} = 0.999$ and $u_{\min} = 0$.
The shifted timesteps used at inference are
\begin{equation}
\tau_j = \frac{s \cdot u_j}{1 + (s - 1) \cdot u_j},
\qquad s = 5,
\label{eq:shifted_t}
\end{equation}
with the endpoints clamped to $\tau_0 = 0.999$ and $\tau_N = 0$; the terminal node $\tau_N = \tau_{50} = 0$ is the clean endpoint reached by the final ODE step rather than a model-call node.
At $s{=}1$ the shift reduces to the identity; with $s{=}5$ the schedule allocates more steps to the high-noise regime where the lip-sync trajectory structure of Sec.~\ref{sec:method_analysis} unfolds.
Table~\ref{tab:shift_mapping} reports representative $j \leftrightarrow \tau_j$ checkpoints.
In particular, the two-step inference schedule $J_{LF}{=}(0, 30)$ corresponds to $\tau \in \{0.999, 0.769\}$, and the windowed-CFG band $j \in [20, 40]$ used by SW-DMD (Sec.~\ref{sec:method_swdmd}) corresponds to $\tau \in [0.555, 0.882]$.

\begin{table}[h]
    \centering
    \caption{\textbf{Shifted-ODE schedule with shift $s{=}5$ and $N{=}50$ steps.} Step indices $j$ map to shifted timesteps $\tau_j$ via Eqs.~\ref{eq:uniform_t}--\ref{eq:shifted_t}; representative rows used as checkpoints throughout the paper are reported.}
    \label{tab:shift_mapping}
    \vspace{4pt}
    \begin{tabular}{c c c}
    \toprule
    Step $j$ & $u_j$ (uniform) & $\tau_j$ (shifted) \\
    \midrule
    0  & 0.999 & 0.999 \\
    5  & 0.899 & 0.978 \\
    10 & 0.799 & 0.952 \\
    15 & 0.699 & 0.921 \\
    20 & 0.599 & 0.882 \\
    25 & 0.499 & 0.833 \\
    30 & 0.399 & 0.769 \\
    35 & 0.299 & 0.681 \\
    40 & 0.199 & 0.555 \\
    45 & 0.099 & 0.357 \\
    49 & 0.019 & 0.093 \\
    \bottomrule
    \end{tabular}
\end{table}

\paragrapht{Metrics.}
SSIM~\cite{wang2004image} is computed with the standard $7{\times}7$ Gaussian window, and LPIPS~\cite{zhang2018unreasonable} uses the standard \texttt{lpips} library implementation with the AlexNet backbone.
Sync-C and Sync-D are computed with the standard SyncNet~\cite{chung2016out} port and use the customary $\pm 15$ frame search window for the optimal audio-visual offset.
All metrics are evaluated on the mouth region defined by the lip-region mask of Sec.~\ref{sup:teacher_finetune}.

\subsection{Full 4-metric trajectory analysis}
\label{sup:trajectory_full}

\begin{figure}[h]
    \centering
    \includegraphics[width=\linewidth]{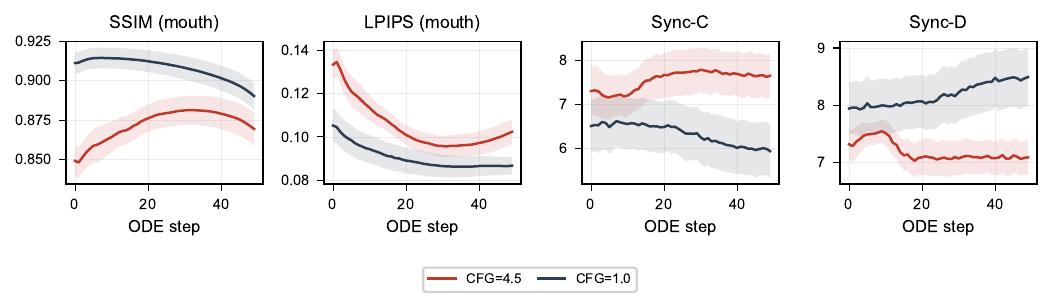}
    \caption{\textbf{The CFG fidelity--sync tradeoff (full 4-metric).} Per-step mean across $n{=}10$ samples; shaded bands are $\pm 1$ standard error. Red: CFG-guided teacher ($s{=}4.5$); navy: no-CFG teacher ($s{=}1.0$). SSIM (mouth) tracks LPIPS, and Sync-D mirrors Sync-C: the same separation between the two trajectories observed in the main figure is reproduced on these additional metrics.}
    \label{fig:cfg_tradeoff_full}
\end{figure}

\begin{figure}[h]
    \centering
    \includegraphics[width=\linewidth]{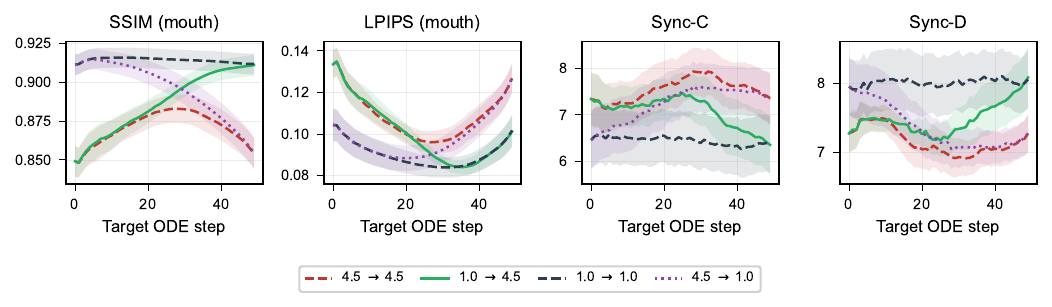}
    \caption{\textbf{Euler-step $2\times 2$ factorial (full 4-metric).} Per-step mean across $n{=}10$ samples; shaded bands are $\pm 1$ standard error. Each trace is one cell of $(s_0, s_1)$. The reference-axis pattern (cells sharing $s_0$ converge by mid-trajectory) holds on SSIM as well as LPIPS; the sync-axis pattern (single-CFG cells close most of the gap to CFG$\to$CFG around step 30, then diverge outside the mid-trajectory window) holds on Sync-D as well as Sync-C.}
    \label{fig:cfg_factorial_full}
\end{figure}

\begin{figure}[h]
    \centering
    \includegraphics[width=\linewidth]{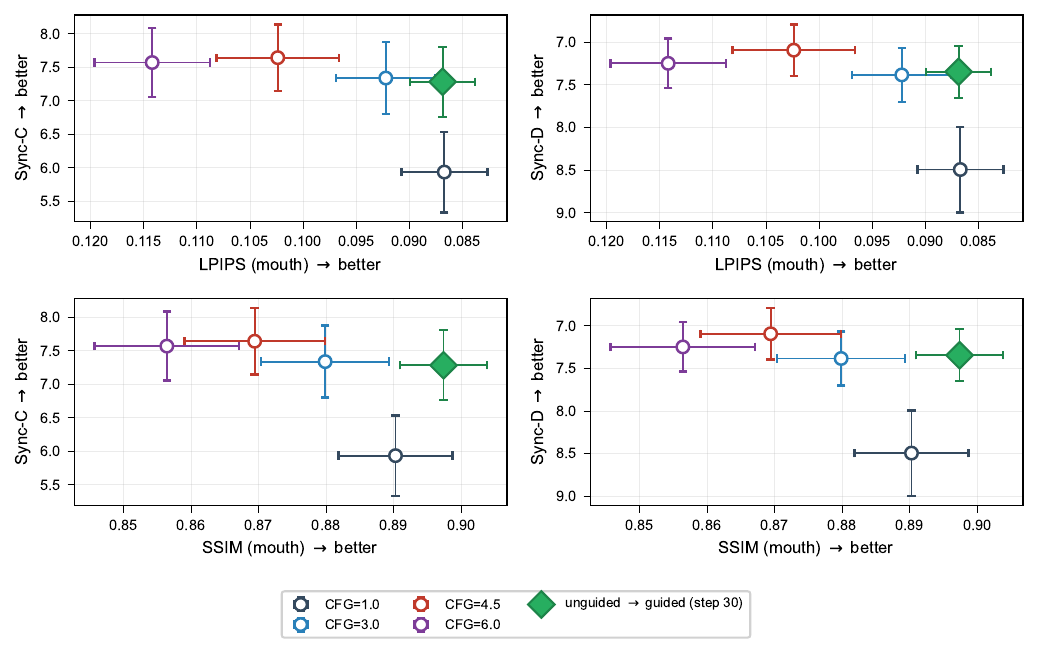}
    \caption{\textbf{Fixed-CFG endpoints vs. schedule operating point (full 4-panel).} Step-49 endpoints of fixed-CFG sweeps at $s\in\{1.0, 3.0, 4.5, 6.0\}$ as open circles; the no-CFG$\to$CFG Euler-step operating point at $j{=}30$ as a filled green diamond. Both axes carry $\pm 1$~SE error bars on $n{=}10$ samples. Axes are oriented so up-right is favorable (LPIPS, Sync-D inverted). The Sync-D panels (the right column) tell the same story as the Sync-C panels reproduced in the main paper.}
    \label{fig:frontier_full}
\end{figure}

The main paper (Fig.~\ref{fig:analysis_combined}) shows LPIPS (mouth) and Sync-C, the two metrics that carry the CFG fidelity--sync tradeoff and the schedule decomposition story.
The full 4-metric versions (adding SSIM (mouth) on the reference side and Sync-D on the sync side) replicate the same patterns: the CFG fidelity--sync tradeoff (Fig.~\ref{fig:cfg_tradeoff_full}), the Euler-step $2\times 2$ factorial (Fig.~\ref{fig:cfg_factorial_full}), and the fixed-CFG frontier comparison against the schedule operating point (Fig.~\ref{fig:frontier_full}).

\subsection{Audio-only CFG variant cross-checks}
\label{sup:audio_only_drop}
\begin{figure}[h]
    \centering
    \includegraphics[width=\linewidth]{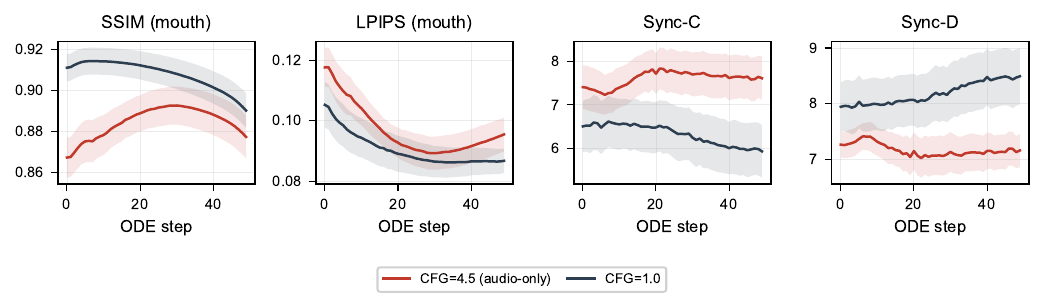}
    \caption{\textbf{CFG fidelity--sync tradeoff, audio-only drop mode.} Audio-only counterpart of main Fig.~\ref{fig:analysis_combined}(a) (full 4-metric counterpart in Fig.~\ref{fig:cfg_tradeoff_full}). Per-step means on the mouth region across $n{=}10$ samples; shaded bands are $\pm 1$ standard error. Red: $s{=}4.5$ with audio-only drop. Navy: $s{=}1.0$ (drop mode irrelevant when guidance scale is 1.0). The same direction of separation holds across all four metrics under audio-only drop.}
    \label{fig:cfg_tradeoff_audio_only}
\end{figure}

The main paper's trajectory analysis (Sec.~\ref{sec:method_analysis}) uses OmniAvatar's standard text+audio CFG drop mode (`cfg\_drop\_text=true').
We re-run the same trajectory variants at `cfg\_drop\_text=false' (audio-only drop) as a sanity check that the result holds under the alternate CFG drop mode.
Per-step means on the mouth region match the text+audio trajectory within sub-$0.2$ Sync-C points across all variants and within $0.02$ on SSIM and LPIPS (Fig.~\ref{fig:cfg_tradeoff_audio_only}), confirming that the CFG fidelity--sync tradeoff and its trajectory structure are not artifacts of dropping text.

\begin{figure}[h]
    \centering
    \includegraphics[width=\linewidth]{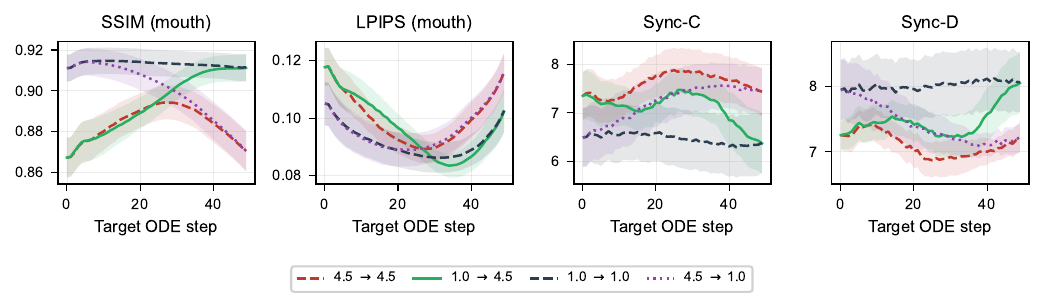}
    \caption{\textbf{Euler-step CFG factorial, audio-only drop mode.} Audio-only counterpart of main Fig.~\ref{fig:analysis_combined}(b) (full 4-metric counterpart in Fig.~\ref{fig:cfg_factorial_full}). Per-step means on the mouth region across $n{=}10$ samples; shaded bands are $\pm 1$ standard error. Same four cells $(s_0, s_1)$ as the main paper: $s_0$ drives the velocity from noise; $s_1$ is used at the re-evaluated landing. Both axes of separation persist: cells sharing $s_0$ converge on SSIM and LPIPS by mid-trajectory; both single-CFG cells (green, purple) close most of the sync gap to the CFG\,$\to$\,CFG ceiling around landings near step~30, and diverge outside this window.}
    \label{fig:cfg_factorial_audio_only}
\end{figure}

The Euler-step $2\times 2$ factorial (Sec.~\ref{sec:method_analysis_factorial}) likewise replicates under audio-only drop (Fig.~\ref{fig:cfg_factorial_audio_only}).

\subsection{Trajectory plateau details}
\label{sup:plateau}
\begin{figure}[h]
    \centering
    \includegraphics[width=\linewidth]{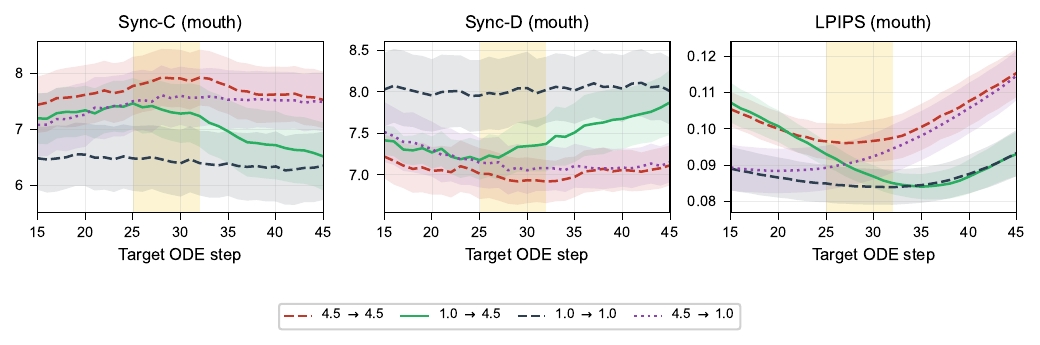}
    \caption{\textbf{Trajectory plateau zoom around the joint reference-sync optimum.} Per-step means on the mouth region across $n{=}10$ samples; shaded bands are $\pm 1$ standard error. Same four Euler-step cells $(s_0, s_1)$ as main Fig.~\ref{fig:analysis_combined}(b), restricted to landing steps $j_1 \in [15, 45]$. The plateau region $j_1 \in [25, 32]$ is shaded gold. Within the plateau, the no-CFG\,$\to$\,CFG cell (green) achieves Sync-C and Sync-D close to the CFG\,$\to$\,CFG ceiling (red) while keeping mouth-LPIPS close to the no-CFG\,$\to$\,no-CFG floor (navy) --- the recipe's joint optimum.}
    \label{fig:plateau_zoom}
\end{figure}

Zooming in on the Euler-step landing-step axis around the §4.3 plateau (Fig.~\ref{fig:plateau_zoom}), the no-CFG\,$\to$\,CFG cell sustains Sync-C, Sync-D, and mouth-LPIPS within standard error across landing steps $j_1 \in [25, 32]$.
A paired $t$-test on Sync-D between $j_1{=}25$ and $j_1{=}30$ on the no-CFG\,$\to$\,CFG cell does not reject equality ($t{=}{-}1.69$, $p{=}0.13$, $n{=}10$), so the plateau is statistically flat across this window.
We therefore treat any landing step in this range as a valid representative; the trajectory analysis in §4 uses $j_1{=}30$.

\section{Method (extended)}
\label{sup:method_extended}

\subsection{Hyperparameters and training details}
\label{sup:hyperparams}

We provide here additional training and inference details that supplement Sec.~\ref{sec:exp_settings}.

\paragrapht{Optimization and schedule.}
Both stages use AdamW with weight decay $0.01$ and gradient norm clipped to $10.0$, run in \texttt{bf16} mixed precision on 4 NVIDIA H200 GPUs at an effective batch size of $64$ via gradient accumulation.
Stage~1 (Diffusion Forcing pretraining; Sec.~\ref{sec:method_overview}) runs for $5\mathrm{K}$ steps at learning rate $10^{-5}$ with a $1\mathrm{K}$-step linear warmup and the standard AdamW betas $(0.9, 0.999)$.
Stage~2 (Self Forcing DMD distillation with our recipe) runs for $600$ steps at learning rate $2{\times}10^{-6}$ for both the student and the fake-score critic with no warmup; we set $\beta_1{=}0$ on both networks to disable momentum across their alternating updates, retaining $\beta_2{=}0.999$.
The student-to-fake-score update ratio is $5{:}1$, following existing pipelines~\cite{yin2024improved,yin2024one,huang2025self}.

\paragrapht{Stage-specific timestep sampling and fake-score initialization.}
Both stages share the filtered training dataset of Sec.~\ref{sup:teacher_data}.

Stage~1 supervises the student under a block-wise inhomogeneous timestep schedule: each three-latent-frame chunk is noised at one timestep $\tau_j$ drawn uniformly from the discrete shifted-ODE grid $\{\tau_j\}_{j \in J_{LF}}$ matching the inference schedule used at that training setting (e.g., $\{\tau_0, \tau_{30}\}$ at $J_{LF}{=}(0,30)$), so chunks are noised independently across the rollout while frames within a chunk share their noise level.

Stage~2 (Self Forcing DMD distillation) operates on the student's self causal rollout.
At each iteration, the student performs a chunk-wise causal denoising rollout through its $K{=}2$-call few-step schedule $J_{LF}{=}(0,30)$, producing the clean prediction $\hat{X}_\theta = \{\hat{x}_0^i\}_{i=1}^{N}$ that aggregates per-chunk outputs (Algorithm~\ref{alg:lipforcing}).
This $\hat{X}_\theta$ is the object DMD supervises: we draw a continuous timestep $t \sim q(t)$ on the shifted range $[0.001, 0.999]$ (shift~$5$, Eq.~\ref{eq:shifted_t}; the distribution $q(t)$ introduced in Sec.~\ref{sec:preliminaries}) and re-noise as $x_t = (1{-}t)\,\hat{X}_\theta + t\,\epsilon$ with fresh noise $\epsilon \sim \mathcal{N}(0, I)$.
Both the teacher and the fake-score critic evaluate at this $x_t$.
Crucially, $t$ is sampled independently of the student's rollout timesteps: the rollout traces the discrete few-step schedule $\{\tau_j\}_{j \in J_{LF}}$, while DMD supervision randomizes $t$ over the full continuous range.
The windowed schedule $s_{\mathrm{SW}}$ of Eq.~\ref{eq:swdmd} applies as defined in Sec.~\ref{sec:method_swdmd}: the teacher is queried with CFG when $j(t) \in [20, 40]$ (shifted-$t$ band $[0.555, 0.882]$) and without CFG outside it.
The fake-score critic is initialized from the OmniAvatar~\cite{gan2025omniavatar} release weights at the matching student scale and is trained without classifier-free guidance at any timestep.

\paragrapht{Architecture, rollout, and inference.}
The 1.3B student is trained with full finetuning; the 14B student uses LoRA~\cite{hu2022lora} at rank $128$, $\alpha{=}64$ on attention and FFN layers following the OmniAvatar~\cite{gan2025omniavatar} convention, with the audio-conditioning projections and the patch embedding fully finetuned alongside.
Causal AR rollout proceeds in chunks of three latent frames; the first frame is held fixed as an attention sink, and the context window covers the six most recent frames outside the sink.
At inference, the student takes $K{=}2$ denoising steps per chunk at $J_{LF}{=}(0,30)$ (Sec.~\ref{sec:method_inference}) with no classifier-free guidance; the windowed schedule of Eq.~\ref{eq:swdmd} applies during distillation only.

\paragrapht{Total compute.}
All training and distillation runs use NVIDIA H200 GPUs ($2$ GPUs for the $1.3$B teacher fine-tune per Sec.~\ref{sup:teacher_training}; $4$ GPUs for every other run).
Teacher fine-tuning (Sec.~\ref{sup:teacher_finetune}) takes $\sim$$3$ days at $1.3$B and $\sim$$1$ week at $14$B ($\sim$$140$ and $\sim$$670$ H200-hours respectively, $\sim$$810$ H200-hours combined).
Stage~1 Diffusion Forcing pretraining takes $33$\,h at $1.3$B and $42$\,h at $14$B ($\sim$$300$ H200-hours combined).
Stage~2 DMD distillation takes $13$\,h per $1.3$B run and $17$\,h for the $14$B run; after deduplicating the eleven distinct $1.3$B ablation cells across Tabs.~\ref{tab:components}--\ref{tab:second_step} and the headline $14$B run, Stage~2 totals $\sim$$640$ H200-hours.
Inference and evaluation across all benchmarks add roughly $5\%$ of the training cost ($\sim$$90$ H200-hours).
Reported runs in this paper therefore account for $\sim$$1{,}900$ H200-hours; including preliminary experiments, hyperparameter searches, and design iterations not reported, total project compute is estimated at approximately $2{\times}$ this figure ($\sim$$3{,}800$ H200-hours).

\subsection{SyncNet reward implementation details}
\label{sup:syncnet_reward_details}

The SyncNet reward (Eq.~\ref{eq:reward}) takes the form of a per-sample multiplicative weight $w(\hat{x}_0) = \exp(\beta \cdot R(D(\hat{x}_0), \mathbf{a}))$ on the DMD generator gradient (Eq.~\ref{eq:dmd_update}), where $R$ is the raw SyncNet confidence between the conditioning audio $\mathbf{a}$ and the visual input decoded from the student's clean prediction $\hat{x}_0$.
Both the parameterization and the strength $\beta{=}2$ are inherited from Re-DMD~\cite{lu2025reward} without modification.
We use the lightweight Tiny AutoEncoder (TAE)~\cite{BoerBohan2025TAEHV} for the decoder $D$ rather than the Wan 3D VAE, so that the per-step reward forward pass adds minimal latency and memory overhead alongside the teacher, fake-score critic, and SyncNet model already resident on the GPU.

The reward weight is forward-only: gradients flow through $\partial \hat{x}_0 / \partial \theta$ in Eq.~\ref{eq:reward} only, with the SyncNet model, the TAE decoder, and the audio embedding pathway all detached from the backward pass.
SyncNet is the standard port also used for the trajectory analysis (Sec.~\ref{sup:analysis_setup_details}); the decoded frames are resized to SyncNet's expected input dimensions without any further mouth-region cropping, since the training clips are already face-aligned at preprocessing (Sec.~\ref{sup:teacher_data}).
Audio is processed with SyncNet's conventional mel-feature pipeline.

\subsection{Streaming rollout details}
\label{sup:streaming_details}

Causal AR rollout proceeds chunk by chunk: each chunk denoises three latent frames in parallel and attends causally to past frames in a rolling KV cache, plus a held-fixed attention sink at temporal position~0.
After Wan 3D VAE decoding the first chunk produces nine pixel frames (the VAE's leading latent is $1{\times}$-compressed rather than $4{\times}$-compressed) and each subsequent chunk produces twelve.
The mask is block-causal with sink size~$1$ and a six-frame rolling window outside the sink, following the Self Forcing convention~\cite{huang2025self}; the sink mechanism itself is the long-rollout-stabilizing structure used by recent Self Forcing-derived video diffusion models~\cite{shin2025motionstreamrealtimevideogeneration,yang2025longlive,liu2025rollingforcingautoregressivelong,yi2025deep} to anchor identity across extended rollouts.

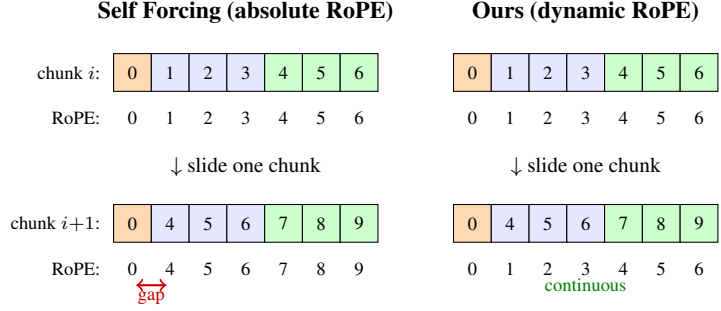
\begin{figure}[t]
\centering
\begin{tikzpicture}[
    box/.style={draw, rectangle, minimum width=0.5cm, minimum height=0.5cm, font=\scriptsize, inner sep=1pt},
    sink/.style={box, fill=orange!30},
    win/.style={box, fill=blue!10},
    cur/.style={box, fill=green!20},
    rope/.style={font=\scriptsize, anchor=center},
    ttl/.style={font=\small\bfseries, anchor=center},
    sub/.style={font=\scriptsize, anchor=east},
]

\node[ttl] at (1.75, 3.0) {Self Forcing (absolute RoPE)};

\node[sub] at (-0.05, 2.2) {chunk $i$:};
\node[sink] at (0.25, 2.2) {0};
\node[win]  at (0.75, 2.2) {1};
\node[win]  at (1.25, 2.2) {2};
\node[win]  at (1.75, 2.2) {3};
\node[cur]  at (2.25, 2.2) {4};
\node[cur]  at (2.75, 2.2) {5};
\node[cur]  at (3.25, 2.2) {6};

\node[sub] at (-0.05, 1.6) {RoPE:};
\node[rope] at (0.25, 1.6) {0};
\node[rope] at (0.75, 1.6) {1};
\node[rope] at (1.25, 1.6) {2};
\node[rope] at (1.75, 1.6) {3};
\node[rope] at (2.25, 1.6) {4};
\node[rope] at (2.75, 1.6) {5};
\node[rope] at (3.25, 1.6) {6};

\node[font=\footnotesize] at (1.75, 0.95) {$\downarrow$ slide one chunk};

\node[sub] at (-0.05, 0.2) {chunk $i{+}1$:};
\node[sink] at (0.25, 0.2) {0};
\node[win]  at (0.75, 0.2) {4};
\node[win]  at (1.25, 0.2) {5};
\node[win]  at (1.75, 0.2) {6};
\node[cur]  at (2.25, 0.2) {7};
\node[cur]  at (2.75, 0.2) {8};
\node[cur]  at (3.25, 0.2) {9};

\node[sub] at (-0.05, -0.4) {RoPE:};
\node[rope] at (0.25, -0.4) {0};
\node[rope] at (0.75, -0.4) {4};
\node[rope] at (1.25, -0.4) {5};
\node[rope] at (1.75, -0.4) {6};
\node[rope] at (2.25, -0.4) {7};
\node[rope] at (2.75, -0.4) {8};
\node[rope] at (3.25, -0.4) {9};

\draw[<->, thick, red!80!black] (0.30, -0.62) -- (0.70, -0.62);
\node[font=\scriptsize, red!80!black] at (0.50, -0.78) {gap};

\begin{scope}[shift={(4.5, 0)}]
\node[ttl] at (1.75, 3.0) {Ours (dynamic RoPE)};

\node[sink] at (0.25, 2.2) {0};
\node[win]  at (0.75, 2.2) {1};
\node[win]  at (1.25, 2.2) {2};
\node[win]  at (1.75, 2.2) {3};
\node[cur]  at (2.25, 2.2) {4};
\node[cur]  at (2.75, 2.2) {5};
\node[cur]  at (3.25, 2.2) {6};
\node[rope] at (0.25, 1.6) {0};
\node[rope] at (0.75, 1.6) {1};
\node[rope] at (1.25, 1.6) {2};
\node[rope] at (1.75, 1.6) {3};
\node[rope] at (2.25, 1.6) {4};
\node[rope] at (2.75, 1.6) {5};
\node[rope] at (3.25, 1.6) {6};

\node[font=\footnotesize] at (1.75, 0.95) {$\downarrow$ slide one chunk};

\node[sink] at (0.25, 0.2) {0};
\node[win]  at (0.75, 0.2) {4};
\node[win]  at (1.25, 0.2) {5};
\node[win]  at (1.75, 0.2) {6};
\node[cur]  at (2.25, 0.2) {7};
\node[cur]  at (2.75, 0.2) {8};
\node[cur]  at (3.25, 0.2) {9};
\node[rope] at (0.25, -0.4) {0};
\node[rope] at (0.75, -0.4) {1};
\node[rope] at (1.25, -0.4) {2};
\node[rope] at (1.75, -0.4) {3};
\node[rope] at (2.25, -0.4) {4};
\node[rope] at (2.75, -0.4) {5};
\node[rope] at (3.25, -0.4) {6};

\node[font=\scriptsize, green!50!black] at (1.75, -0.62) {continuous};
\end{scope}

\end{tikzpicture}
\caption{\textbf{Streaming attention sink and dynamic RoPE.} Cache state across two consecutive chunks under our streaming setup: $1$-frame sink plus a $6$-frame rolling window comprising one cached past block of $3$ frames and the current $3$-frame chunk being denoised, for a total cache size of $7$ frames. Boxes are colored by region (orange = sink, blue = cached past block, green = current chunk being denoised); numbers inside boxes are absolute frame indices in the rollout, numbers below are the temporal RoPE indices the model receives. \textbf{Without} dynamic RoPE (left), the RoPE index equals the absolute frame index, so the sink-to-window gap grows by one chunk every transition (here from $0{\to}1$ at chunk $i$ to $0{\to}4$ at chunk $i{+}1$), driving positional inputs out of the training distribution as rollouts extend. \textbf{With} dynamic RoPE (right, after MotionStream~\cite{shin2025motionstreamrealtimevideogeneration}), RoPE indices are assigned by cache slot, so the same positions $0{\ldots}6$ are presented to the model regardless of rollout length.}
\label{fig:dynamic_rope}
\end{figure}

The KV cache is rolling and never recomputed: at each chunk transition the oldest non-sink entries are evicted to make room for the incoming chunk.
Each chunk runs three forward passes through the model: the two denoising steps at $J_{LF}{=}(0,30)$, plus a final pass on the resulting clean latent that writes its KVs into the cache as causal context for subsequent chunks.
Following MotionStream~\cite{shin2025motionstreamrealtimevideogeneration}, we cache the pre-RoPE KVs and apply temporal RoPE according to each entry's position within the cache rather than its absolute index in the rollout (Fig.~\ref{fig:dynamic_rope}); this keeps the sink-to-window gap fixed at the training-time value, so the model never sees positional offsets larger than those encountered during pretraining as the rollout extends to long horizons.
Audio is processed for the entire input sequence by the OmniAvatar Audio Pack (Sec.~\ref{sup:omniavatar_overview}) and then sliced per chunk; the Audio Pack already aligns audio embeddings with the latent temporal axis, so per-chunk slicing is a direct index into the pre-extracted features.

\subsection{Algorithm pseudocode}
\label{sup:algorithm_extended}

Algorithm~\ref{alg:lipforcing} gives the full \modelname training iteration.
The structure follows the Self Forcing training algorithm~\cite{huang2025self} extended with chunk-wise audio conditioning as in Live Avatar~\cite{huang2025liveavatarstreamingrealtime}, plus three modifications introduced in this paper: (i) the windowed teacher schedule $s_{\mathrm{SW}}$ from SW-DMD (Sec.~\ref{sec:method_swdmd}, Eq.~\ref{eq:swdmd}), (ii) the SyncNet reward weighting $w$ on the generator gradient (Sec.~\ref{sec:method_reward}, Eq.~\ref{eq:reward}), and (iii) the analysis-derived two-step student schedule $J_{LF}{=}(0,30)$ over which the supervision call is sampled (Sec.~\ref{sec:method_inference}).
Fake-score critic updates follow the standard DMD pipeline at the student-to-fake-score ratio of Sec.~\ref{sup:hyperparams} and are omitted from the algorithm for brevity.

\begin{algorithm}[h]
\setstretch{1.1}
\caption{\modelname training iteration.}
\label{alg:lipforcing}
\begin{algorithmic}[1]
\Require Student call indices $J_{LF}{=}(0, 30)$; let $\tau_{j'}$ denote the timestep that follows $\tau_j$ in $J_{LF}$
\Require Number of chunks $N$; chunk-wise conditioning $c_{1:N}$ (Sec.~\ref{sup:teacher_finetune}); audio $\mathbf{a}$
\Require Generator $G_\theta$ with KV-returning variant $G_\theta^{\mathrm{KV}}$; fake-score critic $\phi$
\Require Frozen teacher providing $S_{\mathrm{real}}$; frozen TAE decoder $D$; frozen SyncNet $\mathrm{Sync}$
\Require Reward strength $\beta{=}2$; fake-score update period $K_{\mathrm{fs}}{=}5$
\Loop
    \State Initialize student rollout $\hat{\mathbf{X}}_\theta \leftarrow []$; KV cache $\mathrm{KV} \leftarrow []$
    \State Sample supervision call index $j^\star \sim \mathrm{Unif}(J_{LF})$ \Comment{$j^\star\in\{0,30\}$}
    \For{$i = 1, \ldots, N$}
        \State Initialize $x^i_{\tau_0} \sim \mathcal{N}(0, I)$
        \For{$j \in J_{LF}$ in order, until $j = j^\star$}
            \If{$j = j^\star$}
                \State Enable gradient computation
                \State $\hat{x}_0^i \leftarrow G_\theta(x^i_{\tau_j};\, \tau_j,\, \mathrm{KV},\, c_i)$
                \State $\hat{\mathbf{X}}_\theta\mathrm{.append}(\hat{x}_0^i)$
                \State Disable gradient computation
                \State $\mathrm{kv}^i \leftarrow G_\theta^{\mathrm{KV}}(\hat{x}_0^i;\, 0,\, \mathrm{KV},\, c_i)$ \Comment{clean-latent KV, as at inference}
                \State $\mathrm{KV}\mathrm{.append}(\mathrm{kv}^i)$
            \Else
                \State Disable gradient computation
                \State $\hat{x}_0^i \leftarrow G_\theta(x^i_{\tau_j};\, \tau_j,\, \mathrm{KV},\, c_i)$
                \State Sample $\epsilon \sim \mathcal{N}(0, I)$
                \State $x^i_{\tau_{j'}} \leftarrow (1{-}\tau_{j'})\hat{x}_0^i + \tau_{j'}\epsilon$
            \EndIf
        \EndFor
    \EndFor
    \State Sample DMD timestep $t \sim q(t)$, noise $\epsilon \sim \mathcal{N}(0, I)$
    \State Re-noise for DMD: $x_t \leftarrow (1{-}t)\hat{\mathbf{X}}_\theta + t\epsilon$
    \State Look up CFG scale: $s_t \leftarrow s_{\mathrm{SW}}(j(t))$ \Comment{SW-DMD, Eq.~\ref{eq:swdmd}}
    \State Compute teacher score $S^{\mathrm{CFG}}_{\mathrm{real}}(x_t,\, t,\, c;\, s_t)$ \Comment{Eq.~\ref{eq:cfg}}
    \State Compute fake score $S_{\mathrm{fake}}(x_t,\, t,\, c)$
    \State Reward weight: $w \leftarrow \mathrm{stop\_grad}\!\left(\exp\!\big(\beta \cdot \mathrm{Sync}(D(\hat{\mathbf{X}}_\theta), \mathbf{a})\big)\right)$ \Comment{SyncNet reward, Eq.~\ref{eq:reward}}
    \State Update $\theta$ via $w \cdot \nabla_\theta \mathcal{L}_{\mathrm{DMD}}$ \Comment{Eq.~\ref{eq:dmd_update}}
\EndLoop
\end{algorithmic}
\end{algorithm}

\section{Experiments (extended)}
\label{sup:experiments_extended}

\subsection{Compute and efficiency methodology}
\label{sup:compute_methodology}

Efficiency measurements run on a single NVIDIA H100 80\,GB GPU.
Reported throughput and time-to-first-frame are measured from the first VAE encode to the end of the first chunk's last VAE decode, so audio preprocessing, face detection, and any post-decode compositing or paste-back fall outside the timing window; the same convention is applied uniformly to every baseline to keep the comparison fair.
Each baseline is run from its publicly released code and checkpoint (Wav2Lip~\cite{Prajwal_2020}, VideoReTalking~\cite{cheng2022videoretalkingaudiobasedlipsynchronization}, Diff2Lip~\cite{mukhopadhyay2023diff2lipaudioconditioneddiffusion}, MuseTalk~\cite{zhang2025musetalkrealtimehighfidelityvideo}, LatentSync~\cite{li2024latentsync}, and X-Dub~\cite{he2025inpainting}) at default inference settings; we make no architectural or weight modifications to any baseline.

\begin{figure}[h]
  \centering
  \includegraphics[width=0.5\linewidth]{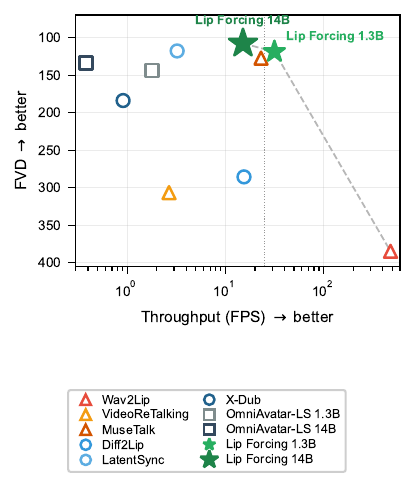}
  \caption{\textbf{Throughput--FVD Pareto frontier across all baselines on HDTF.} Companion to the diffusion-only chart in the main paper (Fig.~\ref{fig:teaser}): adds the single-pass methods Wav2Lip~\cite{Prajwal_2020}, VideoReTalking~\cite{cheng2022videoretalkingaudiobasedlipsynchronization}, and MuseTalk~\cite{zhang2025musetalkrealtimehighfidelityvideo} that are excluded from the main-body diffusion-only comparison. Self Forcing and the ground-truth row are still omitted; the FVD axis is inverted so the up-right corner is the best Pareto position. Vertical dotted line: $25$-FPS playback rate; dashed line: Pareto frontier. \modelname (14B) achieves the lowest FVD on the chart ($107.88$), while Wav2Lip's frontier position is throughput-only -- its FVD ($384.82$) is $\sim 3.5\times$ that of \modelname (14B), which is why the main-paper figure restricts the comparison to diffusion-based peers.}
  \label{fig:pareto_all}
\end{figure}

Figure~\ref{fig:pareto_all} visualizes the throughput--quality tradeoff across the full baseline set on a log-FPS axis.
\modelname (1.3B, 14B) sit on the meaningful Pareto frontier; the only other frontier point is Wav2Lip~\cite{Prajwal_2020}, whose throughput-only frontier position comes at the cost of an FVD penalty of $\sim 3.5\times$ over \modelname (14B), which is why the main-paper Pareto figure restricts the comparison to diffusion-based methods.

The 1.3B variant at $512{\times}512$ with two-step inference, \texttt{torch.compile}, and TAE decoding reaches 31.58\,FPS, comfortably above the 25\,FPS playback rate of the test videos.
Causal sliding-window attention bounds the DiT KV cache at sink\,$+$\,window\,$=$\,$1{+}6{=}7$ latent frames regardless of rollout length, so peak GPU memory in the streaming-plus-TAE configuration is 8.78\,/\,9.39\,GB (allocated\,/\,reserved) at 1.3B and 40.63\,/\,41.38\,GB at 14B; switching the decoder to the Wan~VAE adds about 1.5\,GB on top of each figure, and X-Dub sits between the two scales at 18.60\,/\,19.28\,GB.

\subsection{User study full protocol}
\label{sup:user_study_protocol}

We run a self-hosted Mean Opinion Score (MOS) user study on a 30-clip pool sourced from HDTF~\cite{zhang2021flow} and TalkVid~\cite{chen2025talkvidlargescalediversifieddataset}, comparing \modelname (14B) against six baselines.
Each rater views ten sample pages, with five HDTF clips and five TalkVid clips drawn at random; every page shows the ground-truth video alongside three model outputs anonymized as A, B, and C with the method-to-label assignment randomized per page.
For each anonymized output, raters provide four 5-point Likert ratings covering Video and Audio Synchronization, Video Quality, ID Preservation, and Naturalness, yielding 30 model evaluations per rater across the four axes.

\subsection{Datasets and additional benchmarks}
\label{sup:dataset_details}

We evaluate \modelname on three test sets that probe complementary axes of generalization beyond the main-paper HDTF\textsubscript{short} comparison.
All test clips pass through the alignment pipeline of Sec.~\ref{sup:teacher_data} (face detection, InsightFace~\cite{deng2019arcface} affine alignment, and a $512{\times}512$ crop); after generation, we invert the affine transform to paste the synthesized face region back into the original frame before computing metrics.

\paragrapht{HDTF\textsubscript{short} and HDTF\textsubscript{long}.}
We follow the dual-setting protocol of recent work~\cite{li2024latentsync}: HDTF\textsubscript{short} drives the main-paper comparison on 33 81-frame clips matching our training configuration, and HDTF\textsubscript{long} extends to full-length videos up to 6 minutes to stress long-horizon temporal stability under streaming rollout.
HDTF\textsubscript{long} numbers and a cross-identity audio variant on HDTF\textsubscript{short} are reported in Sec.~\ref{sup:long_video_extended} and Sec.~\ref{sup:cross_audio_extended}.

\paragrapht{Hallo3.}
For Hallo3~\cite{cui2025hallo3highlydynamicrealistic} we hold out 30 videos chosen at the start of the project and never seen during training, providing an out-of-domain check against a more dynamic talking-head distribution.
Per-method results appear in Table~\ref{tab:hallo3}.

\begin{table}[h]
  \centering
  \caption{\textbf{Hallo3 evaluation (30 held-out clips).} Quality, identity, and sync metrics across baselines and \modelname. Best values \textbf{bold}; second-best \underline{underlined}.}
\label{tab:hallo3}
\begin{tabular}{l rr cccc}
\toprule
Method & FVD $\downarrow$ & FID $\downarrow$ & SSIM $\uparrow$ & CSIM $\uparrow$ & Sync-D $\downarrow$ & Sync-C $\uparrow$ \\
\midrule
Wav2Lip~\cite{Prajwal_2020}                                                 & 271.55             & 19.70            & 0.9262             & 0.9411             & \underline{6.60} & \textbf{8.70}    \\
VideoReTalking~\cite{cheng2022videoretalkingaudiobasedlipsynchronization}    & 190.21             & 21.36            & 0.9011             & 0.8996             & 6.94             & 7.83             \\
MuseTalk~\cite{zhang2025musetalkrealtimehighfidelityvideo}                   & 136.16             & 8.88             & 0.9317             & 0.9234             & 8.38             & 6.17             \\
\midrule
Diff2Lip~\cite{mukhopadhyay2023diff2lipaudioconditioneddiffusion}            & 178.64             & 20.11            & 0.9217             & 0.9321             & \textbf{6.22}    & 8.26             \\
LatentSync~\cite{li2024latentsync}                                           & 109.21             & \textbf{6.84}    & \underline{0.9443} & \underline{0.9424} & 6.71             & \underline{8.38} \\
X-Dub~\cite{he2025inpainting}                                                & 199.85             & 13.67            & 0.8518             & 0.8792             & 7.79             & 7.47             \\
\midrule
\modelname (1.3B)                                                            & \underline{101.25} & 7.83             & 0.9321             & 0.9300             & 8.78             & 5.96             \\
\modelname (14B)                                                             & \textbf{87.85}     & \underline{7.12} & \textbf{0.9482}    & \textbf{0.9464}    & 8.23             & 6.58             \\
\bottomrule
\end{tabular}

\end{table}

\paragrapht{TalkVid.}
For TalkVid we hold out 30 self-driven clips ($1920{\times}1080$, 25\,fps, $\sim$3\,s each) chosen at the start of the project and never seen during training, with audio and video drawn from the same source clip.
Per-method results appear in Table~\ref{tab:talkvid}.

\begin{table}[h]
  \centering
  \caption{\textbf{TalkVid evaluation (self-driven, 30 held-out clips).} Quality, identity, and sync metrics across baselines and \modelname. Best values \textbf{bold}; second-best \underline{underlined}.}
\label{tab:talkvid}
\begin{tabular}{l rr cccc}
\toprule
Method & FVD $\downarrow$ & FID $\downarrow$ & SSIM $\uparrow$ & CSIM $\uparrow$ & Sync-D $\downarrow$ & Sync-C $\uparrow$ \\
\midrule
Wav2Lip~\cite{Prajwal_2020}                                                 & 382.64             & 50.28            & 0.7777             & 0.9553             & 7.14             & 8.11             \\
VideoReTalking~\cite{cheng2022videoretalkingaudiobasedlipsynchronization}    & 318.22             & 54.61            & 0.7242             & 0.9301             & 7.20             & 7.74             \\
MuseTalk~\cite{zhang2025musetalkrealtimehighfidelityvideo}                   & 294.84             & 29.46            & 0.7312             & 0.9497             & 8.89             & 5.77             \\
\midrule
Diff2Lip~\cite{mukhopadhyay2023diff2lipaudioconditioneddiffusion}            & 286.86             & 46.47            & 0.7689             & 0.9585             & \textbf{6.45}    & \underline{8.41} \\
LatentSync~\cite{li2024latentsync}                                           & 171.96             & 25.39            & 0.7928             & \underline{0.9629} & \underline{6.75} & \textbf{8.78}    \\
X-Dub~\cite{he2025inpainting}                                                & 191.23             & 14.62            & 0.8202             & 0.9245             & 7.90             & 7.63             \\
\midrule
\modelname (1.3B)                                                            & \underline{118.32} & \underline{9.17} & \underline{0.9095} & 0.9542             & 8.53             & 6.29             \\
\modelname (14B)                                                             & \textbf{111.98}    & \textbf{8.79}    & \textbf{0.9300}    & \textbf{0.9649}    & 7.69             & 7.24             \\
\bottomrule
\end{tabular}

\end{table}

\subsection{Long-video evaluation extended}
\label{sup:long_video_extended}

\begin{figure}[t]
  \centering
  \includegraphics[width=\linewidth,keepaspectratio]{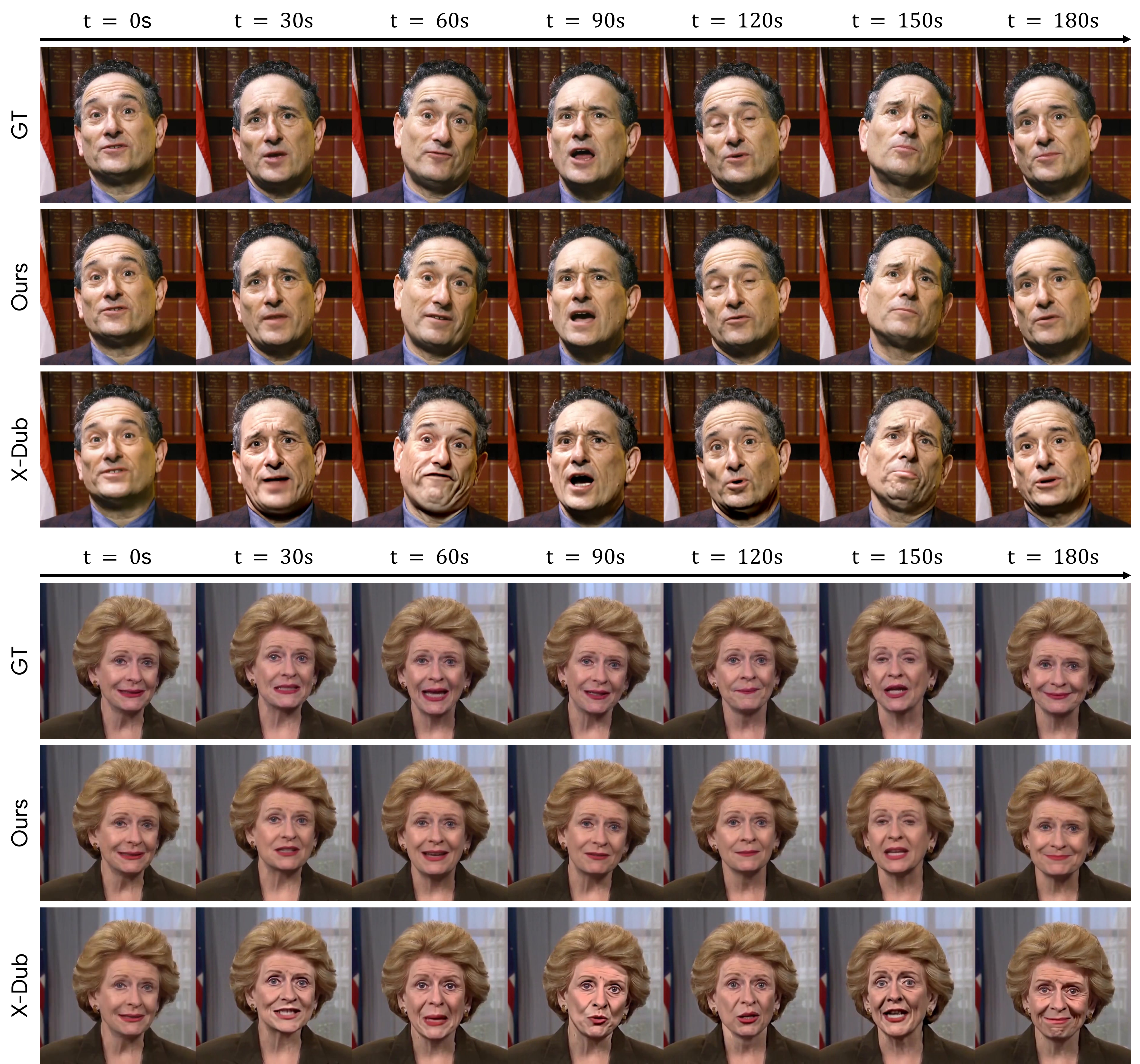}
  \caption{\textbf{Long-video qualitative results on HDTF\textsubscript{long}.}
  Two identities, each rolled out to $t{=}180$\,s and sampled every $30$\,s, comparing ground truth, \modelname, and the strongest baseline X-Dub at consistent timestamps.
  Frame quality, identity, and background remain stable across the full $3$-minute rollout under \modelname's causal AR streaming, well beyond the $81$-frame ($\sim$$3.24$\,s) training chunk.}
  \label{fig:qual_long_video}
\end{figure}

We evaluate \modelname against the same baseline set on long videos from the HDTF test set, where causal AR rollout must hold quality, sync, and temporal consistency over horizons well beyond the 81-frame training chunk (Table~\ref{tab:long_video}).
Qualitative samples appear in Fig.~\ref{fig:qual_long_video}.
We find that our method is robust to error accumulation, and is able to stably generate long sequences far beyond our training horizon. 
On the other hand, segment-wise extrapolation done by X-Dub~\cite{he2025inpainting} quickly begins to produce artifacts like over-saturation or identity drift. 

\begin{table}[h]
\centering
\small
\caption{\textbf{Long-video evaluation on HDTF.} Quality, identity, and sync metrics on HDTF\textsubscript{long} clips up to 6 minutes in duration. Best values \textbf{bold}; second-best \underline{underlined}.}
\label{tab:long_video}
\begin{tabular}{l rr cccc}
\toprule
Method & FVD $\downarrow$ & FID $\downarrow$ & SSIM $\uparrow$ & CSIM $\uparrow$ & Sync-D $\downarrow$ & Sync-C $\uparrow$ \\
\midrule
Wav2Lip~\cite{Prajwal_2020}                                                 & 372.47             & 20.26            & 0.9045             & 0.9434             & \underline{6.56} & \underline{9.17} \\
VideoReTalking~\cite{cheng2022videoretalkingaudiobasedlipsynchronization}    & 313.40             & 21.53            & 0.8711             & 0.9057             & 6.72             & 8.70             \\
MuseTalk~\cite{zhang2025musetalkrealtimehighfidelityvideo}                   & 143.20             & 6.89             & \underline{0.9390} & \textbf{0.9583}    & 6.87             & 8.50             \\
\midrule
Diff2Lip~\cite{mukhopadhyay2023diff2lipaudioconditioneddiffusion}            & 311.98             & 18.32            & 0.7285             & 0.8749             & 6.61             & 8.67             \\
LatentSync~\cite{li2024latentsync}                                           & 150.51             & \textbf{4.04}    & \textbf{0.9399}    & --                 & \textbf{5.92}    & \textbf{9.98}    \\
\midrule
\modelname (1.3B)                                                            & \underline{133.41} & 23.98            & 0.8797             & 0.9035             & 7.87             & 7.53             \\
\modelname (14B)                                                             & \textbf{118.97}    & \underline{4.27} & 0.9216             & \underline{0.9450} & 6.86             & 8.68             \\
\bottomrule
\end{tabular}

\end{table}

\subsection{Cross-identity evaluation extended}
\label{sup:cross_audio_extended}

\begin{figure}[t]
  \centering
  \includegraphics[width=\linewidth,keepaspectratio]{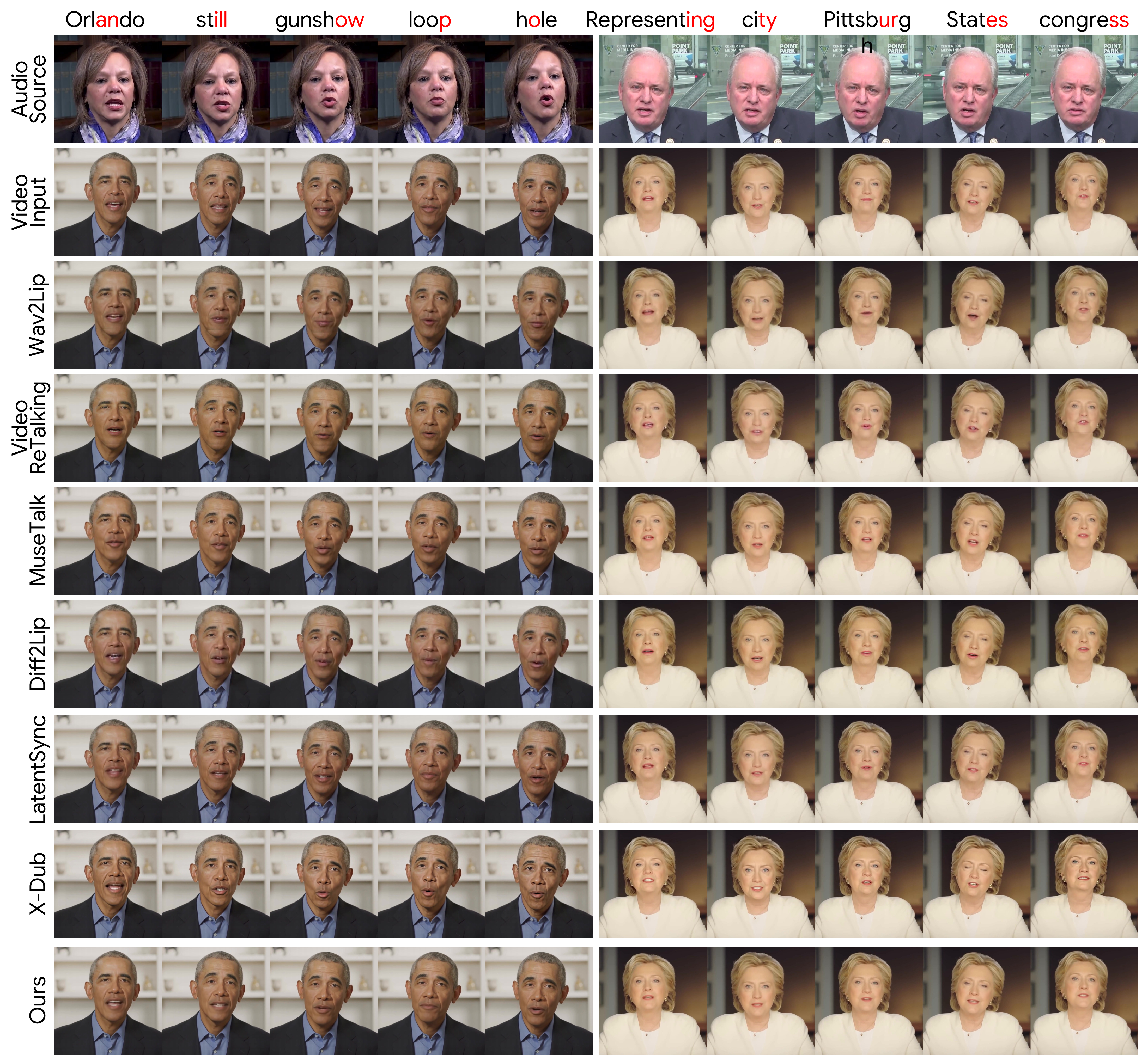}
  \caption{\textbf{Cross-identity qualitative results on HDTF.}
  Two source clips are driven by audio from a different speaker (top row, \emph{Audio Source}); columns mark the moments at which the highlighted English phoneme is articulated.
  Each column compares Wav2Lip, VideoReTalking, Diff2Lip, X-Dub, MuseTalk, LatentSync, and \modelname against the same source frame.
  Lip motion in \modelname follows the driving audio rather than tracking the source speaker's original mouth shape.}
  \label{fig:qual_cross_audio}
\end{figure}

We evaluate \modelname under cross-identity audio drive on HDTF, where the source video is paired with audio from a different speaker (Table~\ref{tab:cross_audio}).
Qualitative samples appear in Fig.~\ref{fig:qual_cross_audio}.
Under cross-identity drive, \modelname preserves the source speaker's identity and visual quality, producing lip motion to the driving audio without identity drift or artifacts.
Consistent with the main comparison (Sec.~\ref{sec:exp_main}), absolute synchronization trails the sync-leaning baselines, and the gap widens under this harder condition where the driving audio is mismatched to the source speaker, reflecting \modelname's fidelity-leaning operating point.

\begin{table}[h]
\centering
\small
\caption{\textbf{Cross-identity evaluation on HDTF.} Source video paired with audio from a different speaker; sync metrics only, since pixel-aligned ground truth does not apply under cross-identity audio. Best values \textbf{bold}; second-best \underline{underlined}.}
\label{tab:cross_audio}
\begin{tabular}{l cc}
\toprule
Method & Sync-C $\uparrow$ & Sync-D $\downarrow$ \\
\midrule
Wav2Lip~\cite{Prajwal_2020}                                                 & 7.75             & 7.54             \\
VideoReTalking~\cite{cheng2022videoretalkingaudiobasedlipsynchronization}    & 7.85             & 6.94             \\
MuseTalk~\cite{zhang2025musetalkrealtimehighfidelityvideo}                   & 6.43             & 8.24             \\
\midrule
Diff2Lip~\cite{mukhopadhyay2023diff2lipaudioconditioneddiffusion}            & \underline{8.30} & \textbf{6.15}    \\
LatentSync~\cite{li2024latentsync}                                           & \textbf{9.05}    & \underline{6.17} \\
X-Dub~\cite{he2025inpainting}                                                & 5.85             & 8.98             \\
\midrule
\modelname (1.3B)                                                            & 5.27             & 9.32             \\
\modelname (14B)                                                             & 6.27             & 8.37             \\
\bottomrule
\end{tabular}

\end{table}

\section{Qualitative Results}
\label{sup:additional_qual}
\begin{figure}[h]
    \centering
    \includegraphics[width=\textwidth]{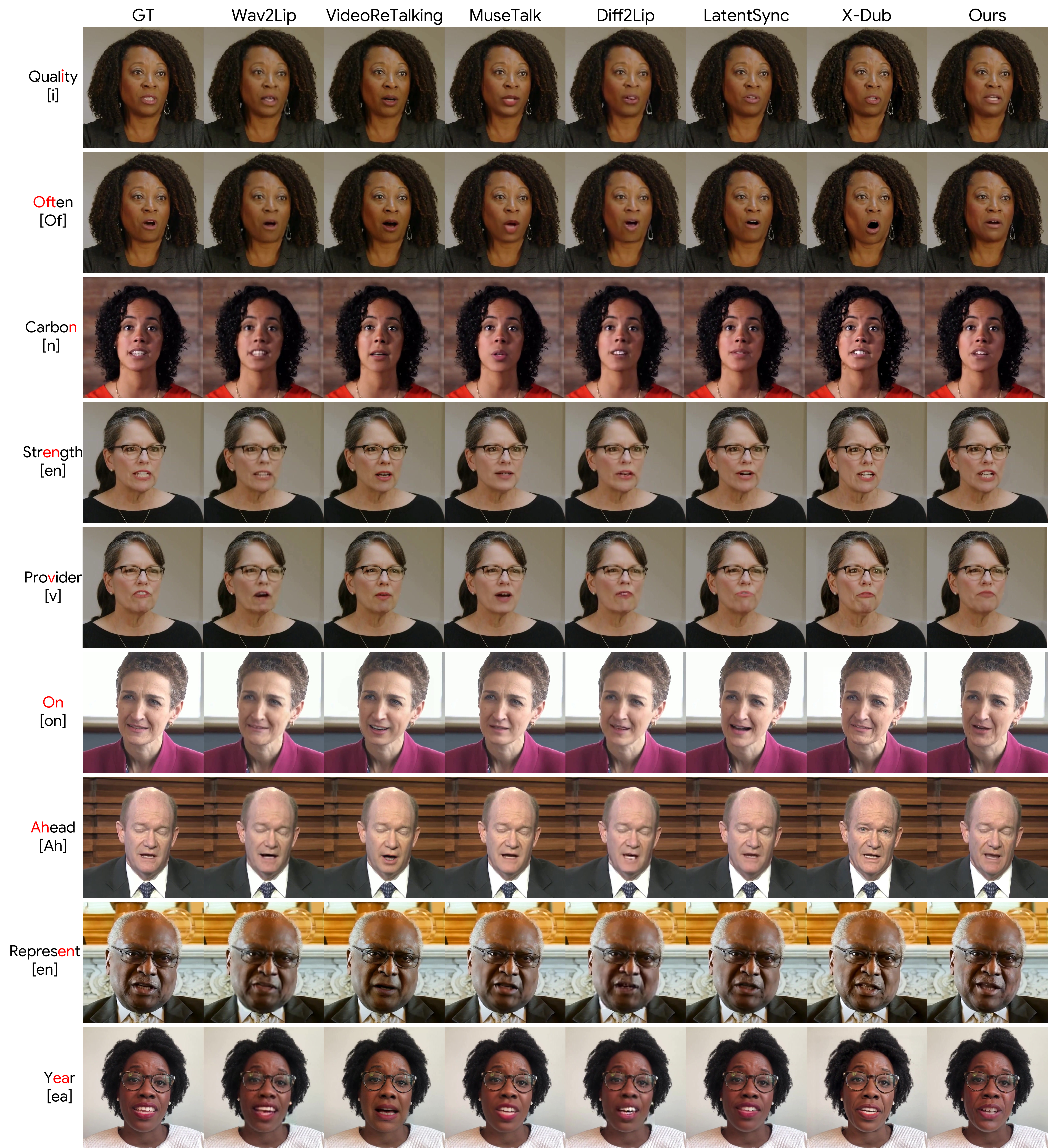}
    \caption{Additional qualitative results from the Hallo3 and HDTF test sets.}
    \label{fig:supple1}
\end{figure}

\begin{figure}[h]
    \centering
    \includegraphics[width=\textwidth]{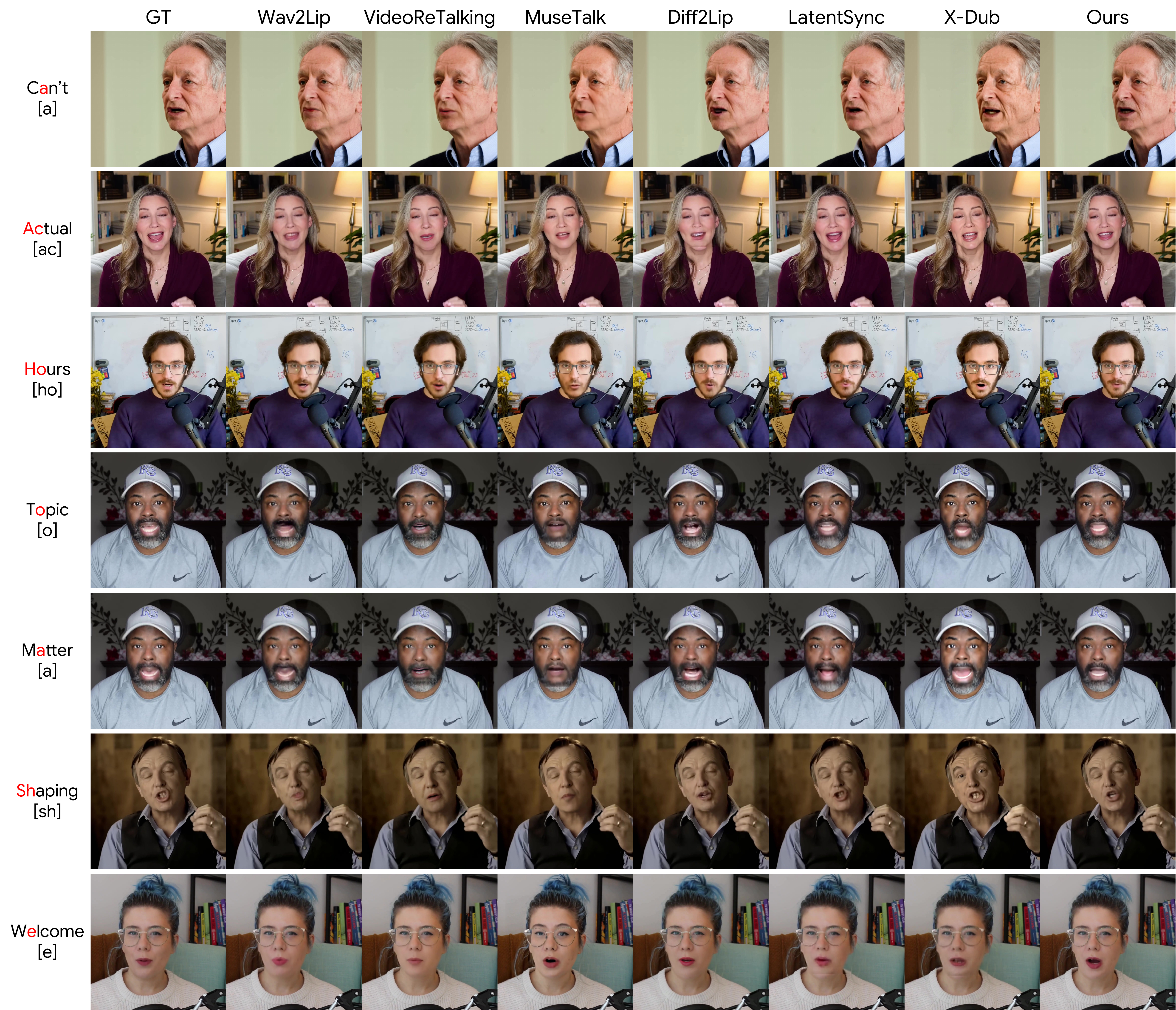}
    \caption{Additional qualitative results from the TalkVid test set.}
    \label{fig:supple2}
\end{figure}

We show additional qualitative results in Fig.~\ref{fig:supple1} and ~\ref{fig:supple2}.

\section{Limitations}
\label{sup:limitations}

\paragrapht{Recipe scope.}
The trajectory-analysis recipe presented in this paper assumes a teacher exhibiting the CFG fidelity--sync tradeoff with a sync-favoring band along the denoising trajectory.
The recipe components --- SW-DMD, the analysis-derived two-step landing, and the SyncNet reward --- target this specific structure.
Teachers without an analogous tradeoff, or whose sync-favoring band lies elsewhere, will not benefit from the windowed schedule directly; the diagnostic methodology of Sec.~\ref{sec:method_analysis} would need to be re-run to identify the appropriate cutoffs.

\paragrapht{Generalization beyond a single teacher lineage.}
The CFG fidelity--sync tradeoff and the schedule cutoffs are characterized on a single 14B OmniAvatar-based teacher, with a robustness check under audio-only CFG drop mode (Sec.~\ref{sup:audio_only_drop}).
We do not claim that this specific trajectory structure or its cutoffs transfer across architectures; the transferable contribution of this paper is the methodology rather than the specific cutoffs.

\paragrapht{Quality gap at 1.3B.}
The 1.3B student is the speed-leading scale and crosses the 25~FPS playback rate of the test videos, but it trails the 14B student on full-frame fidelity.
Applications where lip-sync fidelity is the top priority should adopt the 14B variant; the 1.3B variant is intended for streaming-constrained deployment.

\paragrapht{SyncNet as reward signal.}
Our reward optimizes a SyncNet expert.
Several baselines (Wav2Lip, VideoReTalking) exceed the ground-truth Sync-C, indicating that aggressive optimization of this objective can drift away from perceptual realism.
We mitigate by capping reward strength at $\beta{=}2$ and by reporting both Sync-C and full-frame fidelity metrics, but a more principled audio-visual alignment objective is left for future work.
Because our model's Sync-C remains below the ground-truth value, the capped reward does not appear to induce SyncNet reward-hacking, though the objective remains open to improvement.
\section{Broader Impact}
\label{sup:broader_impact}

\paragrapht{Misuse risks.}
Audio-driven lip synchronization is a dual-use technology.
Beneficial applications include accessibility (live captioning with lip-synced avatars, dubbing for under-served languages), film and game post-production, and human-computer interaction agents.
The same technology can be used to fabricate manipulated video of real individuals --- so-called deepfakes --- with possible consequences for misinformation, fraud, and non-consensual content.

\paragrapht{Real-time amplification.}
The streaming throughput \modelname delivers reduces the marginal cost of generating manipulated content relative to offline pipelines, which is the same property that enables legitimate live applications.
We acknowledge that improvements in efficiency, including those reported in this paper, may lower the barrier to misuse.

\paragrapht{Mitigations.}
We recommend deployments of \modelname be paired with provenance signaling (visible or imperceptible watermarking) and authentication of the user driving the system.
Detection research --- including detectors trained on lip-sync artifacts --- is complementary, and we expect outputs of this and similar systems to enter the training distributions of such detectors.

\paragrapht{Datasets and consent.}
We use four publicly released audio-visual datasets across training and evaluation: VoxCeleb2~\cite{Chung_2018} and Hallo3~\cite{cui2025hallo3highlydynamicrealistic} for training, HDTF~\cite{zhang2021flow} for both training and evaluation, and TalkVid~\cite{chen2025talkvidlargescalediversifieddataset} for evaluation only.
All four datasets are released by their original authors for non-commercial research use; we use them under those terms and have not modified or redistributed the underlying clips.
Users redistributing models trained on these datasets should consult each dataset's license terms (links and license names are listed at each dataset's public release page).

\paragrapht{Baselines and code.}
Lip-sync baselines compared in Sec.~\ref{sec:exp_main} and \supp{sup:experiments_extended} --- Wav2Lip~\cite{Prajwal_2020}, VideoReTalking~\cite{cheng2022videoretalkingaudiobasedlipsynchronization}, Diff2Lip~\cite{mukhopadhyay2023diff2lipaudioconditioneddiffusion}, MuseTalk~\cite{zhang2025musetalkrealtimehighfidelityvideo}, LatentSync~\cite{li2024latentsync}, and X-Dub~\cite{he2025inpainting} --- are run from their publicly released code and checkpoints under the licenses associated with each release; these are predominantly permissive open-source licenses (e.g., MIT, Apache-2.0) for the code, with non-commercial research terms attached to derived weights or datasets where applicable.
Auxiliary tools used in our pipeline (SyncNet~\cite{chung2016out}, ArcFace~\cite{deng2019arcface}, LPIPS~\cite{zhang2018unreasonable}, HyperIQA~\cite{su2020blindly}, the Tiny AutoEncoder~\cite{BoerBohan2025TAEHV}, the Wan~2.1 video diffusion backbone~\cite{wan2025}, and OmniAvatar~\cite{gan2025omniavatar}) are likewise used at their publicly released versions for non-commercial research purposes consistent with each project's stated terms.
Per-asset license names (e.g., CC-BY 4.0 vs. MIT vs. research-only) and version pins will be enumerated in the model card released alongside our code, where any redistribution or derivative-use questions can be resolved against each upstream project's license file.

\clearpage

\end{document}